\documentclass{article}

\PassOptionsToPackage{numbers, compress}{natbib}
 \usepackage[main, final]{neurips_2026}
\usepackage[utf8]{inputenc} % allow utf-8 input
\usepackage[T1]{fontenc}    % use 8-bit T1 fonts
\usepackage{hyperref}       % hyperlinks
\usepackage{url}            % simple URL typesetting
\usepackage{booktabs}       % professional-quality tables
\usepackage{amsmath}        % math environments and \text
\usepackage{amsthm}         % theorem environments
\usepackage{thmtools}       % restatable theorems
\usepackage{amsfonts}       % blackboard math symbols
\usepackage{nicefrac}       % compact symbols for 1/2, etc.
\usepackage{microtype}      % microtypography
\usepackage{xcolor}         % colors
\usepackage{wrapfig}        % wrap figures with text
\usepackage{algorithm}      % floating algorithm environment
\usepackage{algpseudocode}  % algorithmic pseudocode
\usepackage{enumitem}      % custom list spacing

\title{Bridging CLIP with DINO: Cross-Modal Information Maximization for Online Test-Time Adaptation}

\usepackage{multirow}
\usepackage{rotating}
\usepackage{subcaption}

\usepackage{listings}
\usepackage{xcolor}

\definecolor{codegreen}{rgb}{0,0.6,0}
\definecolor{codegray}{rgb}{0.5,0.5,0.5}
\definecolor{codepurple}{rgb}{0.58,0,0.82}

\lstdefinestyle{pytorchstyle}{
    language=Python,
    backgroundcolor=\color{white},
    commentstyle=\color{codegreen},
    keywordstyle=\color{magenta},
    numberstyle=\tiny\color{codegray},
    stringstyle=\color{codepurple},
    basicstyle=\ttfamily\footnotesize, 
    breakatwhitespace=false,
    breaklines=true,
    captionpos=b,
    keepspaces=true,
    numbers=left,
    numbersep=5pt,
    showspaces=false,
    showstringspaces=false,
    showtabs=false,
    tabsize=4
}

\usepackage{xcolor}
\usepackage{colortbl}
\newcommand{\meansd}[2]{#1 $\pm$ #2}

\author{
  Taehoon Kim$^1$\quad
  Wonjun Hwang$^1$\thanks{Corresponding authors}\quad\,
  Hyunsouk Cho$^2$\footnotemark[1] \\
  $^1$Korea University, $^2$Ajou University \\
  \texttt{\{th951113, wjhwang\}@korea.ac.kr, hyunsouk@ajou.ac.kr}%
}

\begin{document}

\maketitle 

\begin{abstract}
Vision-language models such as CLIP enable zero-shot classification from arbitrary class names, but under test-time corruptions, their visual encoder projects distorted features onto incorrect textual anchors with high confidence: a phenomenon we term \emph{modality bias}. Standard entropy minimization reinforces these overconfident errors, which creates a confirmation-bias loop and collapses the predicted label distribution. 
We propose an online test-time adaptation (TTA) method that breaks this loop at three levels. 
At the \emph{representation level}, a frozen self-supervised backbone (e.g., DINO\,v2) is combined with CLIP through a cross-covariance logit ensemble. CLIP text prototypes are projected into the feature space of this text-free backbone, so the self-supervised branch classifies native image features and is less affected by the text bias of CLIP. 
At the \emph{optimization level}, a gated test-time information-maximization objective pairs per-sample sharpening with batch-level diversity, while an adaptive gate excludes unreliable samples from the gradient. At the \emph{prediction level}, an online logit adjustment corrects the residual class-distribution skew.
CDIM outperforms the compared CLIP-based online TTA baselines on ImageNet-C, CIFAR10-C, and CIFAR100-C, and remains competitive in continual and mixed-domain TTA. Extensive analyses and ablation studies verify the contribution of each of its components.
 Our code is available at \href{https://github.com/B1ueC1oud/CDIM}{https://github.com/B1ueC1oud/CDIM}.

\end{abstract}

\section{Introduction}
\label{sec:intro}

\begin{wrapfigure}{r}{0.53\textwidth}
    \vspace{-1ex}
    \centering
    \includegraphics[width=0.5\textwidth]{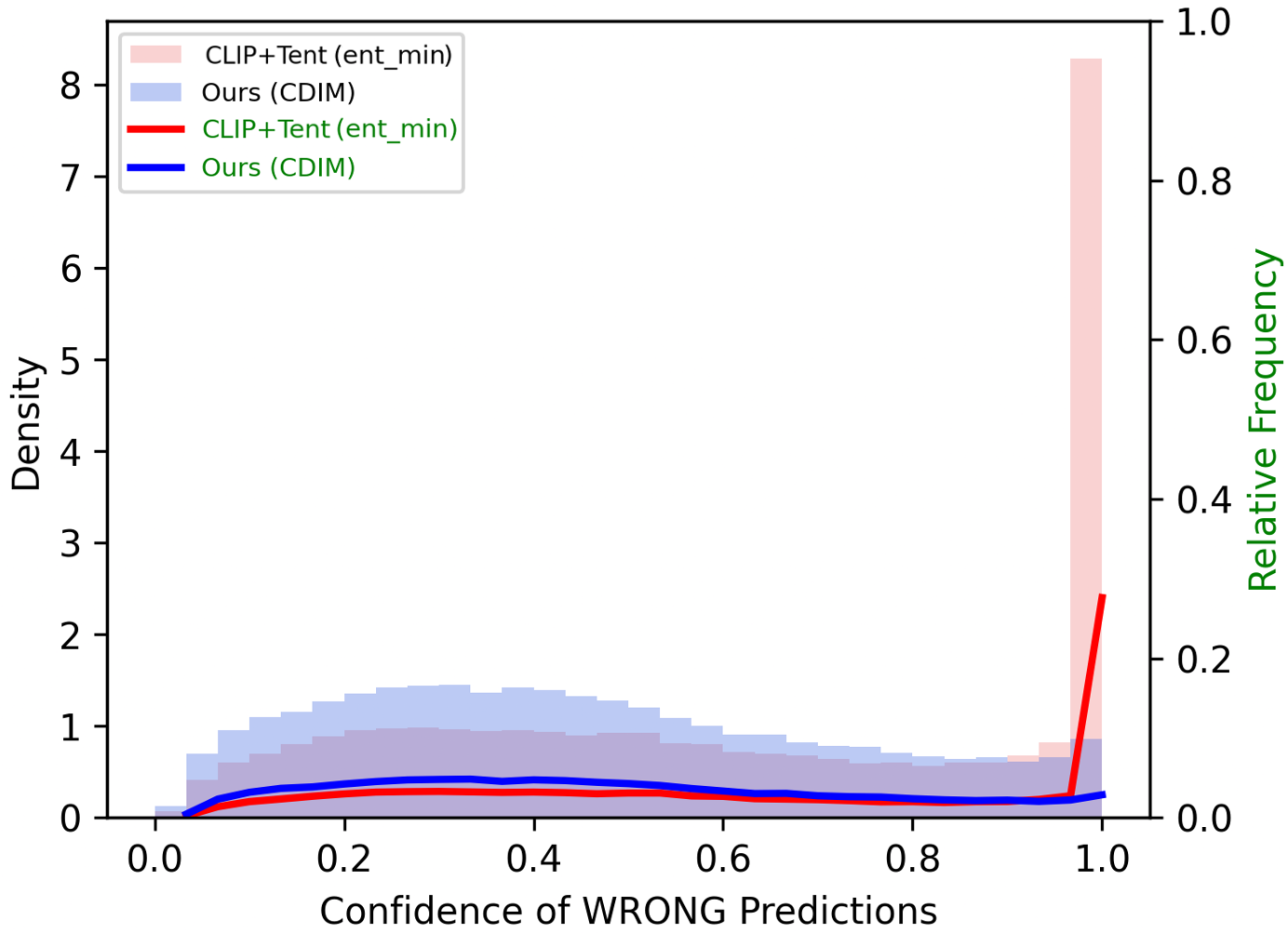}
    \caption{Confidence of incorrect predictions on ImageNet-C (all corruptions, severity 5). CLIP-based entropy minimization (red) assigns high confidence even to incorrect predictions under corruption. CDIM (blue) shifts the confidence of incorrect predictions toward moderate values, which yields a flatter distribution. The corresponding reliability diagram and ECE with 10 equal-width bins are shown in Figure~\ref{fig:calibration}.
    }
    \vspace{-1\baselineskip}
    \label{fig:confidence-hist}
\end{wrapfigure}

The exceptional performance of modern deep learning heavily relies on the premise of identical training and test distributions~\cite{krizhevsky2012imagenet,simonyan2014very,he2016deep}. This premise is routinely violated in practical applications. When deployed in the wild, models often face dynamic environmental changes such as sensor artifacts or weather fluctuations, which lead to severe performance deterioration~\cite{hendrycks2019benchmarking,recht2019imagenet,geirhos2018generalisation,gopalan2011domain,ganin2015unsupervised}.
Fully retraining the model to handle these shifts is often infeasible due to computational constraints or privacy concerns regarding the original source data. 
Test-time adaptation (TTA) addresses this gap by adapting a source-pretrained model during inference, using only a stream of unlabeled target data~\cite{sun2020test, wang2020tent}. 
Because TTA requires neither source data nor a supervised loss, it can be applied in safety-critical settings such as autonomous driving and industrial anomaly detection. 
A recent line of work replaces the source-trained backbone with a vision--language model such as CLIP~\cite{radford2021learning}, removing the cost of task-specific pre-training and providing a zero-shot classifier for any list of class names~\cite{maharana2025batclip,lafon2026cliptta,bao2026mint,zeng2026subspace}. 
However, the open-vocabulary prior of CLIP does not protect the model against the low-level shifts that TTA has to handle: zero-shot accuracy on corrupted inputs~\cite{hendrycks2019benchmarking} drops sharply with corruption severity. Recent analyses suggest that this drop is related to the language supervision of CLIP. The visual encoder of CLIP is strongly aligned with text semantics and projects visual features onto text-interpretable directions~\cite{gandelsman2024interpreting}. As a result, it retains less of the low-level visual structure required for recognition under corruption than text-free self-supervised models do~\cite{oquab2023dinov2,liu2025data}.
When this structure is corrupted, the model tends to project the distorted image onto the nearest, and often incorrect, textual anchor with high confidence \cite{deng2025words}. We refer to this tendency as \emph{modality bias}. Because the model relies on its text bias rather than the corrupted visual evidence, it produces false positives with very high confidence \cite{deng2025words}. 
Entropy-style adapters~\cite{wang2020tent} use the prediction of the model as the learning target, so a confident error is reinforced instead of corrected~\cite{arazo2020pseudo}.
 Modality bias therefore contributes to \emph{confirmation bias} in CLIP-based TTA. The predicted class distribution then drifts away from a balanced one and concentrates on a small subset of anchor classes \cite{lafon2026cliptta}. This is a compound failure: modality bias feeds confirmation bias, which in turn amplifies the class-distribution skew on the next batch.
Figure~\ref{fig:confidence-hist} visualizes the confidence distribution of \emph{incorrect} predictions on ImageNet-C (all corruptions, severity 5). 
CLIP optimized with entropy minimization (TENT, red) concentrates its errors at near-perfect confidence ($>0.95$) during TTA. This pattern is consistent with confirmation bias driven by modality bias: entropy minimization sharpens the class that the model already predicts, so a confident error is reinforced rather than corrected. CLIP+TENT reaches only 25.2\% mean accuracy.

To address this, we propose \textsc{CDIM} (Cross-modal DINO-anchored Information Maximization), which addresses modality bias across three distinct levels of the adaptation pipeline.
First, at the \textbf{representation level}, we augment CLIP with a frozen self-supervised backbone (DINO\,v2 \cite{oquab2023dinov2}) that serves as a complementary anchor, since its training is independent of text supervision.
Recent analyses indicate that CLIP, owing to language supervision, emphasizes high-level semantics, whereas text-free self-supervised models such as DINO are more sensitive to low-level visual structure~\cite{liu2025data,jiang2023clip}. By projecting CLIP's text prototypes into the DINO feature space via a cross-covariance logit ensemble, we stabilize the per-batch prediction against the text-biased drift. 
Second, at the \textbf{optimization level}, we replace standard entropy minimization with gated information maximization. This objective, which we call GT-IM, pairs per-sample sharpening with a batch-level diversity term \cite{krause2010discriminative,liang2020we}. An adaptive confidence gate excludes the most unreliable samples from the sharpening gradient to prevent noisy updates from corrupting the adaptation \cite{niu2022efficient}.
Third, at the \textbf{prediction level}, we correct the residual class-distribution skew at inference time. An online logit adjustment (OLA) tracks the predicted class frequencies with an exponential moving average and counteracts the label-distribution drift that modality bias induces even after the representation has been corrected. OLA follows the logit-adjustment rule of~\cite{menon2020long} and is used here as a uniform-target prior-balancing correction.
Figure~\ref{fig:confidence-hist} (blue) shows that, under CDIM, the confidence of incorrect predictions shifts toward moderate values (0.1 to 0.4).  Incorrect predictions therefore no longer act as near-one-hot targets that entropy-style updates reinforce, which breaks the confirmation-bias loop described above.
\begin{itemize}[leftmargin=1.5em,itemsep=2pt,topsep=2pt]
\item We identify modality bias, the tendency of the CLIP visual encoder to project corrupted features onto incorrect textual anchors with high confidence, as a contributor to confirmation bias in CLIP-based TTA. Figure~\ref{fig:confidence-hist} shows its effect on the confidence distribution.

\item We propose \textsc{CDIM}, which combines a frozen DINO\,v2-anchored logit ensemble, a gated test-time information-maximization objective (GT-IM), and an online logit adjustment to address modality bias at three complementary levels of the adaptation pipeline.
\item Under the online protocol, CDIM outperforms the strongest CLIP-based online TTA baseline by +4.8\%, +3.5\%, and +3.7\% on ImageNet-C, CIFAR10-C, and CIFAR100-C, respectively. It remains competitive under the continual and mixed-domain protocols. These gains require one additional frozen DINO v2 forward pass per batch (Appendix~\ref{app:Resource}).
\end{itemize}

\section{Related Work}

\textbf{TTA with vision–language models.} 
Test-time adaptation (TTA) updates pre-trained models on unlabeled target streams without source data, typically via entropy minimization~\cite{wang2020tent} or teacher-student consistency for continual shifts~\cite{wang2022continual}. 
Recently, TTA has increasingly adopted vision-language models like CLIP~\cite{radford2021learning}. Although early work adapted only the text modality~\cite{shu2022test,feng2023diverse,sui2025just,ma2023swapprompt,pratt2023does,allingham2023simple}, recent methods update the visual or joint encoders using multimodal alignment objectives, such as matching class means or principal directions~\cite{maharana2025batclip, lafon2026cliptta, bao2026mint, zeng2026subspace}. 
Alternatively, \cite{han2025test} adapts self-supervised backbones with mean-feature classifiers to avoid costly supervised pre-training. CDIM instead retains CLIP as the primary backbone and adds a frozen self-supervised model as an auxiliary branch. The representations of both backbones are combined for online adaptation on each test batch.

\textbf{Self-supervised visual backbones next to CLIP.} 
Recent studies show that self-supervised models such as DINO~\cite{caron2021emerging, oquab2023dinov2} extract fine-grained, low-level visual features that complement the high-level semantic representations of CLIP~\cite{jiang2023clip,liu2025data}. Several recent works combine the two models for other tasks:
 NoLA~\cite{imam2024clip} uses DINO to prompt-tune CLIP for label-free classification, while others inject DINO's localization priors to enhance open-vocabulary segmentation~\cite{wysoczanska2024clip,barsellotti2025talking}. Most notably, COSMIC~\cite{huang2025cosmic} integrates CLIP and DINO features in a training-free cache-based protocol with multi-view augmentation.
These works share our intuition that a self-supervised view can recover information lost by the image-text view under shifted or fine-grained conditions. Prior analyses indicate that CLIP mainly encodes high-level semantics, whereas  self-supervised models, exemplified by DINO \cite{oquab2023dinov2, caron2021emerging}, retain more low-level information such as color, style, and fine-grained localization \cite{jiang2023clip, liu2025data}. CDIM combines the two representations to handle the unsupervised domain shifts encountered during TTA.
\begin{figure*}[t]
\centering
\begin{subtable}[t]{0.65\linewidth}
  \centering
  \includegraphics[width=\linewidth]{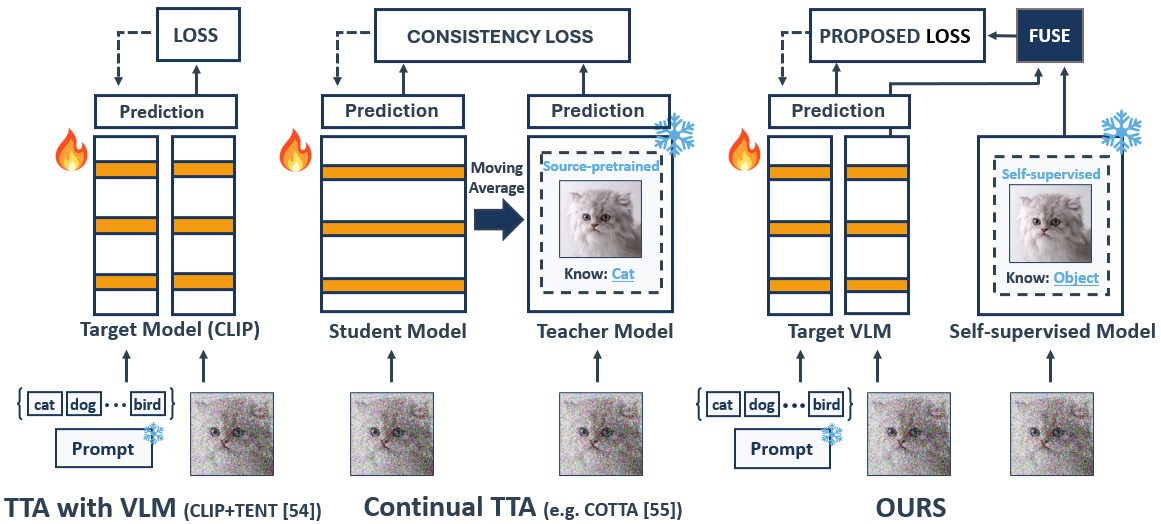}
  \caption{Comparison of \textsc{CDIM} with standard TTA paradigms}
  \label{fig:overall}
\end{subtable}\hfill
\begin{subtable}[t]{0.33\linewidth}
  \centering
  \includegraphics[width=\linewidth,height=4.2cm]{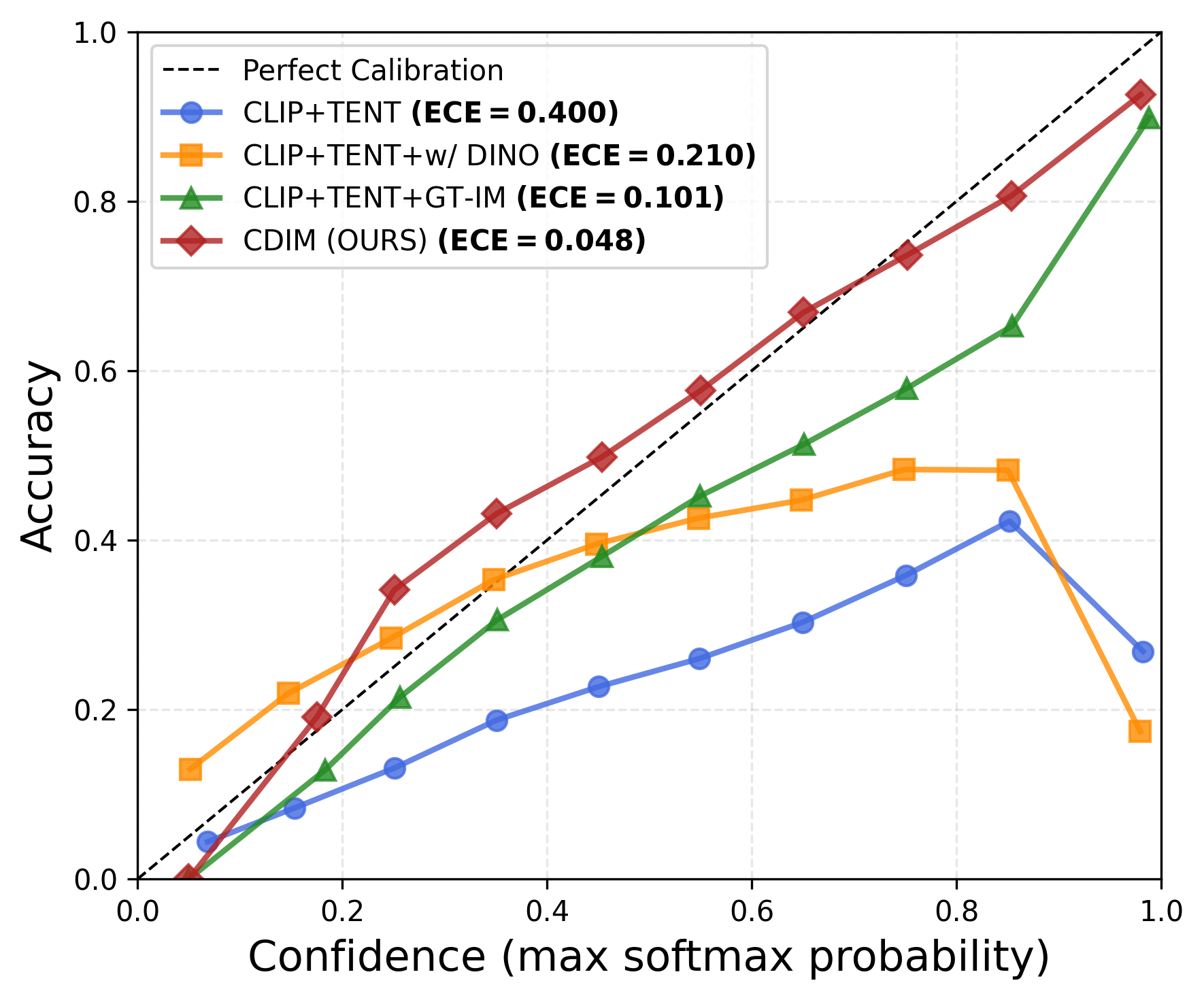}
  \caption{Reliability diagram.}
  \label{fig:calibration}
\end{subtable}
\caption{
      \textcolor{black}{\textbf{(a) }} \underline{Left}: standard TTA uses a vision-language model (e.g., CLIP) that adapts through entropy minimization on its own predictions. \underline{Middle}: continual TTA (CTTA) uses a source-pretrained backbone and a moving-average teacher to enforce teacher-student consistency. \underline{Right}: CDIM adds a frozen self-supervised backbone, in a role similar to that of the teacher in the middle, to support the target CLIP.  \textcolor{black}{\textbf{(b) }} Reliability diagram with 10 equal-width bins. CLIP+TENT (blue) is severely over-confident (below the $y = x$ line, ECE = 0.400). Adding the DINO\,v2 ensemble (orange) or GT-IM (green) alone lowers the ECE to 0.210 and 0.101, respectively, and the full CDIM (red) reaches the lowest ECE (0.048).
      }

\vspace{-3pt}
\end{figure*}

\section{Proposed Method}
\label{sec:method}
 We propose \textsc{CDIM} (\underline{\textbf{C}}ross-modal \underline{\textbf{D}}INO-anchored \underline{\textbf{I}}nformation \underline{\textbf{M}}aximization), an online test-time adaptation method for CLIP-based classifiers. \textsc{CDIM} consists of three core components, each designed to mitigate modality bias at a distinct level of the adaptation pipeline:
\begin{enumerate}[leftmargin=2em,itemsep=1pt,topsep=2pt]
\item \textbf{Representation-level} (Section~\ref{sec:fusion}): a cross-modal logit ensemble with a frozen DINO\,v2 backbone stabilizes the prediction by projecting text prototypes into the DINO feature space, reducing the text-biased drift of CLIP alone.
\item \textbf{Optimization-level} (Section~\ref{sec:tsim}): a gated test-time information-maximization objective (GT-IM) pairs per-sample sharpening with batch-level diversity, while an adaptive confidence gate excludes unreliable samples from the gradient.
\item \textbf{Prediction-level} (Section~\ref{sec:ola}): an online logit adjustment corrects the residual class-distribution skew using an EMA estimate of the class prior.
\end{enumerate}

Figure~\ref{fig:overall} compares CDIM with standard TTA paradigms. Standard TTA with CLIP (left) relies solely on the predictions of the target model through entropy minimization. CTTA (middle) enforces consistency with a moving-average teacher. CDIM (right) uses a frozen self-supervised backbone to anchor the adaptation of the target vision-language model, in a role similar to that of the teacher in CTTA.

% -----------------------------------------------
\subsection{Problem Setup and Notation}
\label{sec:setup}
 
\paragraph{Online TTA with CLIP.}
Following \cite{maharana2025batclip}, we apply the standard online TTA setting~\cite{wang2020tent} specialized to a CLIP-based classifier.
A pre-trained CLIP model $f_\theta = (f^{\text{img}}_\theta, f^{\text{txt}}_\theta)$ is deployed on a sequence of unlabeled test batches $\{\mathbf{X}_t\}_{t=1}^{T}$, each of size $B$, drawn from a target distribution that may itself drift across $t$. 
The model returns its predictions online, and its parameters are updated once per batch by an unsupervised objective. No source data, no target labels, and no second pass over a batch are available~\cite{wang2020tent,wang2022continual}.
The adapter returns at every step a probability vector $p_t(y \mid \mathbf{x}) \in \Delta^{K-1}$ over the $K$ classes named in the prompt template ``a photo of a \texttt{<class>}''. \\
\textbf{Backbones.}
\textsc{CDIM} pairs CLIP with a frozen self-supervised image backbone $g$ from the DINO family~\cite{caron2021emerging,oquab2023dinov2}; we use DINO\,v2 with ViT-B/14 as the default. 
The auxiliary backbone $g$ is frozen throughout the stream and, unlike the weight-averaged teacher of CoTTA \cite{wang2022continual}, is never updated. It provides text-free self-supervised visual features that the contrastive image-text training of CLIP does not directly preserve \cite{liu2025data,jiang2023clip}. Following~\cite{wang2020tent} and~\cite{maharana2025batclip}, the learnable parameters are restricted to the channel-wise affine parameters $\{\gamma_{l,k}, \beta_{l,k}\}$ of the normalization layers in the CLIP image and text encoders. All other parameters of CLIP, together with all parameters of $g$, are frozen.

\textbf{Notation.}
For a batch of $B$ test images, let $\mathbf{C} \in \mathbb{R}^{B \times d_c}$ and $\mathbf{D} \in \mathbb{R}^{B \times d_d}$ denote the $\ell_2$-normalized image features from CLIP and DINO, respectively. Let $\mathbf{T} \in \mathbb{R}^{K \times d_c}$ denote the $\ell_2$-normalized text features for the $K$ class prompts. 
Throughout, $H(p)=-\sum_c p_c\log p_c$ is the Shannon entropy in nats. The inverse temperatures of the CLIP and DINO branches are denoted $\tau_C^{-1}$ and $\tau_D^{-1}$, respectively.

% -----------------------------------------------
\subsection{Representation-level: Cross-Modal Logit Ensemble}
\label{sec:fusion}
Modality bias drives the failure of CLIP under corruption by projecting distorted features toward incorrect textual anchors~\cite{maharana2025batclip,bao2026mint,zeng2026subspace,lafon2026cliptta}. An adapter that relies only on CLIP is therefore limited by its own training signal, because the image-text similarity on which it depends has already been distorted by modality bias.
 To address this, we introduce a frozen, non-contrastive image backbone (DINO\,v2) and combine it with CLIP at the logit level. The resulting ensembled zero-shot classifier is expected to be more robust to the text-biased drift of CLIP alone. \textit{The projection goes from text to DINO space and not the reverse.} The DINO image features $D$ remain in their native feature space and are never projected toward text-interpretable directions, which avoids replicating the text-biased projection that causes modality bias in CLIP (Section~\ref{sec:intro}).

\paragraph{DINO-space text prototypes.}
To enable the frozen DINO\,v2 backbone to act as a zero-shot classifier over the same $K$ classes, we project the CLIP text prototypes into the DINO feature space through the cross-covariance between the image features of the two backbones, computed over the current batch:
\begin{equation}
\label{eq:crosscov}
\bar{\mathbf{M}} \;=\;
\tfrac{1}{B}\, \mathbf{C}^\top \mathbf{D},
\qquad
\widetilde{\mathbf{T}}^{\text{D}} \;=\;
\mathrm{norm}\bigl(\,\bar{\mathbf{M}}^\top \mathbf{T}^\top\bigr) 
\;\in\; \mathbb{R}^{d_d \times K},
\end{equation} 
where $\mathrm{norm}(\cdot)$ denotes column-wise $\ell_2$-normalization. The factor $1/B$ makes $\bar M$ an empirical average whose expectation does not depend on the batch size. The construction is parameter-free and batch-local. It adapts the text prototypes to each test batch without any learned mapping between the two backbones.

Unlike a prior method~\cite{choi2022improving} that builds prototypes from source data before deployment, these prototypes are obtained from the CLIP text features and the current batch alone.

\paragraph{Logit ensemble.} We form the per-backbone logits and their ensemble:
\begin{align}
\label{eq:logits}
\mathbf{Z}^{\text{C}} &= \tau_C^{-1}\,\mathbf{C}\mathbf{T}^\top,
\qquad
\mathbf{Z}^{\text{D}} = \tau_D^{-1}\,\mathbf{D}
                       \widetilde{\mathbf{T}}^{\text{D}},
\\[2pt]
\label{eq:ensemble}
\mathbf{Z}^{\text{ens}} &=
   \tfrac{1}{2}\bigl(\mathbf{Z}^{\text{C}}
                   + \mathbf{Z}^{\text{D}}\bigr),
\qquad
p_{ic} = \mathrm{softmax}_c
        \bigl(\mathbf{Z}^{\text{ens}}_{i,:}\bigr).
\end{align}
Here $\tau_C^{-1}$ is the pretrained logit scale of CLIP, and $\tau_D^{-1}$ is a fixed scale of the DINO branch. The ensembled logits $\mathbf{Z}^{\text{ens}}$ are the quantity on which both the GT-IM objective (Section~\ref{sec:tsim}) and the online logit adjustment (Section~\ref{sec:ola}) operate. 
By averaging the predictions of a text-aligned backbone (CLIP) and a text-free backbone (DINO\,v2), the ensemble is partially robust to the modality bias that affects CLIP alone. On corrupted batches, the DINO-projected classifier classifies native DINO image features and is therefore less directly tied to the text-biased drift of CLIP~\cite{liu2025data,jiang2023clip}. In Figure~\ref{fig:calibration}, the cross-modal ensemble (orange line) lowers the Expected Calibration Error (ECE) of the CLIP+TENT baseline (blue line) from 0.400 to 0.210. Because the underlying objective is unchanged, confirmation bias still persists. The next section therefore introduces an objective that alleviates it.

% -----------------------------------------------
\subsection{Optimization-level: Gated Test-time Information Maximization}
\label{sec:tsim}

The canonical TTA objective is per-sample entropy minimization~\cite{wang2020tent}, which sharpens each prediction toward a single anchor. When applied alone to a contrastively pre-trained classifier such as CLIP, this objective tends to collapse the predicted label distribution onto a reduced subset of the anchors as the stream progresses~\cite{lafon2026cliptta}. This collapse is the test-time form of the confirmation bias identified in pseudo-label training~\cite{arazo2020pseudo,pham2021meta}. The standard mitigation in unsupervised classification pairs the per-sample sharpening with a batch-level diversity term, which yields the information-maximization decomposition~\cite{krause2010discriminative,liang2020we}. This decomposition operationalizes the cluster assumption~\cite{grandvalet2004semi,chapelle2009semi}. We adopt it and add an adaptive confidence gate that filters unreliable samples out of the sharpening term to suit online TTA. The resulting objective is called Gated Test-time Information Maximization (GT-IM) and consists of two parts.

\textbf{Adaptive confidence gating.}
Not all samples in a corrupted batch contribute equally to the adaptation signal.
On severely corrupted inputs, modality bias can distort the ensembled posterior so strongly that the per-sample entropy gradient becomes noise-dominated rather than class-informative. Including such samples in the sharpening term risks propagating noisy gradients into the normalization-layer update~\cite{niu2022efficient}. Because these samples cannot be identified without test labels, the least confident samples in each batch are excluded from the sharpening term instead. Each sample is scored by the negative log-sum-exp of its ensembled logits, which serves as an uncertainty score:
\begin{equation}\label{eq:energy}
E_i = -\log\sum_{c=1}^{K}\exp\bigl(z^{\mathrm{ens}}_{ic}\bigr).
\end{equation}
$E_i$ is low when the largest ensembled logits are large, which we interpret as high confidence.
Because it is computed from the logits rather than from the softmax output, it reflects the magnitude of the logits and is not invariant to a constant shift of $z_i$.
In each batch, the $k = \max(1, \lfloor \rho B \rfloor)$ samples with the largest $E_i$ are discarded from the sharpening term:
\begin{equation}
\label{eq:gate}
\mathcal{G}_i \;=\;
\mathbb{1}\bigl\{E_i \;<\;
   E_{(B-k+1)}\bigr\},
\end{equation}
where $E_{(j)}$ is the $j$-th smallest value in the batch $\{E_1, \dots, E_B\}$. For $B \ge 1/\rho$, this retains $\lceil (1-\rho)B \rceil$ samples. The number of discarded samples is fixed by $\rho$, but the threshold adapts to the score distribution of each batch. 
This contrasts with the absolute entropy threshold of~\cite{niu2022efficient}, which can reject an entire ``hard'' batch. The quantile-based selection consistently retains the most informative $(1{-}\rho)$ fraction, ensuring a stable adaptation signal regardless of the batch difficulty.
 
\textbf{GT-IM objective.}
Let $\mathbf{p}_i = \mathrm{softmax}(\mathbf{Z}^{\text{ens}}_{i,:}) \in \Delta^{K-1}$ be the ensemble posterior on sample $i$, and let $\bar{\mathbf{p}} = (1/B)\sum_{i=1}^{B}\mathbf{p}_i$ be the batch-mean prediction. The GT-IM loss applies the gating mask to the conditional-entropy term while computing the marginal-entropy term from the \emph{full} batch:
\begin{equation}
\label{eq:tsim}
\mathcal{L}_{\text{GT-IM}}
\;=\;
\underbrace{
\frac{\sum_{i=1}^{B}\mathcal{G}_i\,H(\mathbf{p}_i)}
     {\sum_{i=1}^{B}\mathcal{G}_i}
}_{\text{gated conditional entropy}}
\;-\;
\underbrace{H(\bar{\mathbf{p}})}_{\text{marginal entropy}}.
\end{equation}
The gradient of the conditional term sharpens each retained $\mathbf{p}_i$ toward its dominant anchor. The gradient of the marginal term pushes $\bar{\mathbf{p}}$ toward the uniform distribution over the $K$ text anchors, which prevents the label-set collapse observed in~\cite{lafon2026cliptta}. The gate is applied asymmetrically. It excludes the least reliable samples, whose per-sample entropy gradients are noise-dominated, from the sharpening term. The marginal term is computed from the full batch, so that $\bar{\mathbf{p}}$ is not biased by the gate and keeps the retained samples from collapsing onto a few anchors. Eq.~\eqref{eq:tsim} therefore combines two different sample sets, and it is an information-maximization objective rather than the mutual information of a single distribution. The two terms are weighted equally and require no balancing hyperparameter. 
In Figure~\ref{fig:confidence-hist}, \textsc{CDIM} moves the confidence of incorrect predictions from high values  ($>0.95$) to a moderate range ($0.1$--$0.4$), so these predictions no longer serve as near-one-hot targets that the update reinforces. The marginal term additionally penalizes the collapse of the batch-level prediction onto a few classes. With GT-IM the ECE of CLIP+TENT drops from $0.400$ to $0.101$ (Figure~\ref{fig:calibration}).
 
\paragraph{Why GT-IM fits the ensembled-logit setting.}
Two properties of the ensembled CLIP/DINO logits $\mathbf{Z}^{\text{ens}}$ make GT-IM suitable for this setting.
First, the ensembled logits combine two zero-shot classifiers that share the same text-anchor set but read images through two different backbones~\cite{radford2021learning,oquab2023dinov2}.
Their average has a lower ECE than the CLIP-based baseline on corrupted batches (Figure~\ref{fig:calibration}).
The marginal-entropy term of GT-IM acts directly on this averaged posterior, whereas a per-sample entropy update~\cite{wang2020tent} ignores cross-sample information.
Second, the ensembled posterior mixes a trainable backbone (CLIP) with a branch whose image features are frozen (DINO\,v2).
The batch-mean $\bar{\mathbf{p}}$ is therefore partially anchored by a component whose image features are never updated.
The prototypes $\widetilde{\mathbf{T}}^{\text{D}}$ are recomputed for every batch from the current CLIP features and receive no gradient, so the anchor is only partial.
The GT-IM diversity term thus operates on a quantity that is expected to be more stable than the previous CLIP-based prediction on which the label-set collapse of~\cite{lafon2026cliptta} was observed (Figure~\ref{fig:drift}).

\paragraph{Distinctions from prior work.} A prior method~\cite{zhang2022memo} minimizes the entropy of the marginal prediction over augmentations of one test sample. In contrast, the marginal term $H(\bar p)$ in Eq.~\eqref{eq:tsim} is maximized over samples in a batch to prevent label collapse. The information-maximization decomposition follows~\cite{krause2010discriminative,liang2020we}, and~\cite{choi2022improving} uses the same decomposition with source prototypes, which CDIM does not require. GT-IM differs in the asymmetric gate, the marginal term applied to the cross-modal ensembled posterior, and the partial anchoring of $\bar{\mathbf{p}}$ by the frozen DINO\,v2 image features.

% -----------------------------------------------
\subsection{Prediction-level: Online Logit Adjustment}
\label{sec:ola}
Our objective in Eq.~\eqref{eq:tsim} addresses the propagation of modality bias at the \emph{optimization level}: its marginal-entropy term pushes the batch-mean prediction $\bar{\mathbf{p}}$ toward the uniform distribution, preventing the predicted label distribution from collapsing onto a biased subset of text anchors. However, GT-IM operates at the gradient level and acts only on the current batch. It does not correct the \emph{residual class-distribution skew} that has already accumulated in the ensembled logits $\mathbf{Z}^{\text{ens}}$ by the time of the final prediction, which is irreversible in the online TTA setting.  
Residual skew may persist after GT-IM for two reasons.
First, the cross-modal ensemble retains a residual dependence on CLIP, which can cause the model to still favor dominant text anchors~\cite{deng2025words,chen2024quantifying}.
Second, as the model adapts to the online stream~\cite{wang2022continual}, skew inherited from past updates accumulates, and batch-local GT-IM gradients cannot instantaneously correct this prolonged bias.

Both challenges resemble test-time label shift~\cite{park2023label}. Since the residual class skew behaves like a label shift, it can be corrected by a single training-free adjustment of the logits, which we call online logit adjustment (OLA).
OLA is a prior-balancing correction in the spirit of the logit-adjustment rule of~\cite{menon2020long} and the prior-shift correction of~\cite{saerens2002adjusting}. If the posterior of a classifier was formed under a class prior $\pi^{S}$, moving it to a target prior $\pi^{T}$ shifts the logits by
\begin{equation}
f_c^{\mathrm{adj}}(x) = f_c(x) + \log\frac{\pi_c^{T}}{\pi_c^{S}}.
\label{eq:prior_shift}
\end{equation}
We use Eq.~\eqref{eq:prior_shift} with two choices. First, the target is uniform, $\pi_c^{T} = 1/K$. Second, because the class prior implicitly encoded in the ensembled logits under corruption is unknown, $\pi^{S}$ is replaced by the model-induced prior $\bar{\pi}^{(t)}$, a running average of the model's own predicted class frequencies. As $\log \pi_c^{T} = -\log K$ does not change the $\arg\max$, Eq.~\eqref{eq:prior_shift} reduces to
\begin{equation}
z_{ic}^{\mathrm{adj}} = z_{ic}^{\mathrm{ens}} - \log \bar{\pi}_c^{(t)},
\label{eq:ola}
\end{equation}
where $\bar{\pi}_c^{(t)}$ estimates the model-induced (effective source) prior of class $c$ at batch step $t$, not the true target prior.
 Eq.~\eqref{eq:ola} is an anti-frequency penalty. It follows the balanced-error-optimal rule of~\cite{menon2020long} under a uniform target, but it is not a Bayes-optimal estimate of the true label distribution. It presumes that the target class marginal is close to uniform. Figure~\ref{fig:skew} illustrates the predicted class frequencies that this correction acts on.
  The prior is estimated from the model's own predictions via an exponential moving average (EMA) of the per-batch softmax frequencies:

\begin{equation}
\label{eq:ema}
\hat{\pi}_c^{(t)}
\;=\;
\frac{1}{B}\sum_{i=1}^{B} p_{ic},
\qquad
\bar{\pi}_c^{(t)}
\;=\;
\alpha\,\bar{\pi}_c^{(t-1)}
\;+\;
(1-\alpha)\,\hat{\pi}_c^{(t)},
\end{equation}
with $\alpha$ as a hyperparameter and $\bar{\pi}_c^{(0)} = 1/K$ (uniform initialization). 
The EMA smooths out batch-to-batch noise while tracking the class-distribution drift induced by modality bias. The final prediction is $\hat y_i = \arg\max_c z^{\mathrm{adj}}_{ic}$. Consequently, the logit adjustment of Eq.~\eqref{eq:ola} acts as a direct countermeasure to the class-distribution skew that modality bias produces. When the text bias of CLIP causes the model to over-predict an anchor class $c^*$, the EMA prior $\bar\pi^{(t)}_{c^*}$ rises above $1/K$, and the correction $-\log \bar\pi^{(t)}_{c^*}$ lowers the logit of that class. Under-predicted classes receive a positive boost. The correction has no learnable parameters, requires no gradient, and adds negligible computation and memory (a single $K$-dimensional state vector).

\paragraph{Distinctions from prior work.}
Standard logit adjustment addresses class-imbalanced training with known label frequencies~\cite{menon2020long} or extends it to TTA under covariate and label shifts~\cite{park2023label}. 
Our method differs from both because the class-distribution skew does not come from a genuine imbalance in the test data. 
It comes from the \emph{modality bias} of the visual encoder: even on a balanced test set, CLIP under corruption over-predicts certain anchors because its features drift toward the text directions of those anchors~\cite{deng2025words}. OLA corrects this artifact at the prediction level and completes the three-level pipeline with the representation-level (Section~\ref{sec:fusion}) and optimization-level (Section~\ref{sec:tsim}) components.

\paragraph{Calibration (Figure~\ref{fig:calibration}).}
 ECE~\cite{guo2017calibration} is computed with 10 equal-width confidence bins on the maximum softmax probability of $z^{\mathrm{ens}}$, aggregated over all 15 corruptions at severity 5 from a single run. Starting from CLIP+TENT (ECE 0.400), adding the DINO\,v2 ensemble or GT-IM alone lowers ECE to 0.210 and 0.101, respectively, and the full CDIM reaches 0.048. Adding OLA alone to CLIP+TENT yields 0.281 (Figure~\ref{fig:app_abl2_real}), and this reduction partly coincides with an accuracy gain of +5.2\% on ImageNet-C (Table~\ref{tab:abl_ola}). Above a confidence of 0.7, where most predictions fall, the CDIM curve stays close to the diagonal (perfect calibration). It deviates only mildly toward under-confidence between 0.25 and 0.65, and it does not reproduce the over-confidence of CLIP+TENT.

% -----------------------------------------------
\subsection{Adaptation Pipeline}
\label{sec:pipeline}
On each test batch $\mathbf{X}_t$, \textsc{CDIM} executes four steps.
\textbf{(1) Forward pass:} a single pass through CLIP and DINO\,v2 produces $\mathbf{C}$, $\mathbf{D}$, $\mathbf{T}$; the cross-modal logit ensemble of Eq.~\eqref{eq:ensemble} yields $\mathbf{Z}^{\text{ens}}$ and $\mathbf{p}_i$.
\textbf{(2) Adaptive confidence gating:}
the uncertainty score $E_i$ (Eq.~\eqref{eq:energy}) is computed, and the $k=\max(1,\lfloor\rho B\rfloor)$ samples with the largest $E_i$ are masked out via Eq.~\eqref{eq:gate}.
\textbf{(3) Gradient step:}
 a single gradient step on $\mathcal{L}_{\text{GT-IM}}$ (Eq.~\eqref{eq:tsim}) updates only the channel-wise affine parameters of the normalization layers in the image and text encoders of CLIP, with $\tilde{T}_D$ (and hence $Z^{D}$) treated as a constant; DINO\,v2 is never updated.
\textbf{(4) Online logit adjustment:} the EMA class prior $\bar{\pi}_c^{(t)}$ is updated via Eq.~\eqref{eq:ema}, the logits are adjusted via Eq.~\eqref{eq:ola}, and the prediction $\hat{y}_i = \arg\max_c z_{ic}^{\text{adj}}$ is returned.

\begin{table*}[t]
\centering
\scriptsize
\caption{Online TTA accuracy (\%) with ViT-B/16 at severity 5. Top-1 and Top-2 are shown in \textbf{bold} and \underline{underline}, respectively.}
\label{tab:benchmark_all}
\setlength{\tabcolsep}{2pt}
\renewcommand{\arraystretch}{0.9}
\resizebox{0.98\textwidth}{!}{%
\begin{tabular}{@{}ll*{15}{c}c@{}}
\toprule
\cmidrule(lr){3-5}\cmidrule(lr){6-9}\cmidrule(lr){10-13}\cmidrule(lr){14-17}
\multicolumn{2}{l}{Method} 
& \multicolumn{1}{c}{Gauss.} 
& \multicolumn{1}{c}{Shot} 
& \multicolumn{1}{c}{Impul.}
& \multicolumn{1}{c}{Defoc.}
& \multicolumn{1}{c}{Glass}
& \multicolumn{1}{c}{Motion}
& \multicolumn{1}{c}{Zoom}
& \multicolumn{1}{c}{Snow}
& \multicolumn{1}{c}{Frost}
& \multicolumn{1}{c}{Fog}
& \multicolumn{1}{c}{Brit.}
& \multicolumn{1}{c}{Contr.}
& \multicolumn{1}{c}{Elast.}
& \multicolumn{1}{c}{Pixel.}
& \multicolumn{1}{c}{JPEG}
& \multicolumn{1}{c}{Mean $\uparrow$} \\
\midrule
\multirow{9}{*}{\rotatebox{90}{ImageNet-C}} & Zero Shot & 11.1 & 12.5 & 12.1 & 23.3 & 15.3 & 24.4 & 22.7 & 32.4 & 29.9 & 35.9 & 54.1 & 17.3 & 12.7 & 31.0 & 33.3 & 24.5 \\
 & TDA~\cite{karmanov2024efficient} & 12.4 & 14.8 & 14.8 & 23.7 & 16.7 & 26.1 & 24.0 & 33.5 & 32.1 & 38.0 & 55.3 & 19.3 & 14.2 & 33.5 & 35.0 & 26.2 \\
 & TENT~\cite{wang2020tent} & 5.2 & 5.7 & 7.4 & 25.3 & 19.3 & 26.9 & 24.2 & 33.5 & 30.5 & 37.8 & 54.3 & 22.5 & 13.9 & 35.1 & 36.0 & 25.2 \\
 & TPT~\cite{shu2022test} & 8.5 & 9.4 & 10.2 & 24.1 & 15.1 & 25.0 & 24.0 & 34.0 & 32.2 & 37.1 & 55.5 & 16.5 & 13.7 & 33.9 & 33.6 & 24.9 \\
 & WATT~\cite{osowiechi2024watt} & 11.2 & 12.8 & 13.0 & 25.1 & 18.1 & 26.9 & 25.2 & 33.4 & 30.2 & 37.4 & 54.0 & 21.8 & 15.2 & 33.5 & 34.9 & 26.2 \\
 & Mint~\cite{bao2026mint} & 19.7 & 20.4 & 19.4 & \underline{26.9} & 21.5 & 29.9 & 26.0 & 32.7 & 29.4 & 38.8 & 55.1 & 24.3 & 19.1 & 36.4 & \underline{38.2} & 29.2 \\
 & BATCLIP~\cite{maharana2025batclip} & 19.5 & 21.0 & 19.3 & 25.9 & 21.3 & 30.0 & 28.5 & 35.2 & 31.4 & 40.2 & 55.2 & 25.7 & 23.7 & 36.8 & 37.2 & 30.1 \\
 & SubTTA~\cite{zeng2026subspace} & \underline{21.2} & \underline{22.9} & \underline{21.4} & 25.3 & \underline{24.8} & \underline{30.7} & \underline{29.6} & \underline{37.2} & \underline{35.9} & \underline{41.3} & \underline{57.0} & \underline{29.7} & \underline{29.4} & 37.6 & \underline{38.2} & \underline{32.1} \\
  & \textbf{CDIM (Ours)} & \bf{26.0} & \bf{27.9} & \bf{26.7} & \bf{31.1} & \bf{30.1} & \bf{35.9} & \bf{34.4} & \bf{42.2} & \bf{38.3} & \bf{45.6} & \bf{59.6} & \bf{34.1} & \bf{33.1} & \bf{45.1} & \bf{43.8} & \bf{36.9} \\
\midrule
\multirow{9}{*}{\rotatebox{90}{CIFAR10-C}} & Zero Shot & 37.9 & 41.7 & 54.4 & 71.7 & 40.9 & 67.9 & 73.7 & 73.9 & 77.3 & 70.2 & 84.4 & 62.4 & 53.8 & 47.7 & 59.5 & 61.2 \\
 & TDA~\cite{karmanov2024efficient} & 40.9 & 44.3 & 50.5 & 72.8 & 44.4 & 71.6 & 76.0 & 75.3 & 77.8 & 71.8 & 86.1 & 62.3 & 56.3 & 47.8 & 57.7 & 62.4 \\
 & TENT~\cite{wang2020tent} & 15.4 & 18.3 & 38.2 & 81.4 & 21.5 & 76.3 & 82.2 & 83.6 & 82.2 & 80.6 & 89.8 & 80.7 & 63.5 & 58.8 & 56.3 & 61.9 \\
 & TPT~\cite{shu2022test} & 37.8 & 42.2 & 60.7 & 72.8 & 44.8 & 69.7 & 75.4 & 76.0 & 78.9 & 72.2 & 85.7 & 62.0 & 58.9 & 55.2 & 62.6 & 63.6 \\
 & WATT~\cite{osowiechi2024watt} & 46.0 & 53.2 & 60.3 & 75.0 & 38.6 & 71.5 & 75.9 & 77.7 & 80.2 & 76.3 & 87.6 & 75.6 & 55.4 & 62.0 & 63.2 & 66.6 \\
 & Mint~\cite{bao2026mint} & 54.4 & 58.7 & 64.5 & 76.5 & 48.9 & 77.9 & 79.3 & 81.6 & 81.5 & 77.2 & 89.9 & 74.5 & 62.1 & 61.4 & 64.5 & 70.2 \\
 & BATCLIP~\cite{maharana2025batclip} & \underline{61.9} & 65.4 & 67.1 & 80.1 & 55.5 & 80.0 & 81.6 & 82.5 & 83.7 & 80.4 & 88.5 & 81.2 & 69.1 & 62.7 & 67.3 & 73.8 \\
 & SubTTA~\cite{zeng2026subspace} & 60.2 & \underline{66.6} & \underline{69.7} & \underline{81.5} & \underline{59.9} & \underline{81.2} & \underline{82.4} & \underline{83.7} & \underline{83.8} & \underline{83.5} & \underline{91.4} & \underline{86.3} & \underline{70.9} & \underline{74.4} & \underline{69.4} & \underline{76.3} \\
  & \textbf{CDIM (Ours)} & \bf{67.8} & \bf{70.9} & \bf{76.3} & \bf{83.2} & \bf{67.1} & \bf{83.0} & \bf{85.2} & \bf{86.2} & \bf{87.0} & \bf{84.8} & \bf{92.5} & \bf{87.0} & \bf{74.9} & \bf{77.5} & \bf{73.4} & \bf{79.8} \\
\midrule
\multirow{9}{*}{\rotatebox{90}{CIFAR100-C}} & Zero Shot & 19.6 & 21.4 & 25.3 & 42.5 & 20.1 & 43.1 & 47.9 & 48.4 & 49.7 & 41.6 & 57.0 & 34.5 & 29.2 & 23.9 & 32.5 & 35.8 \\
 & TDA~\cite{karmanov2024efficient} & 22.6 & 25.1 & 29.2 & 43.1 & 19.4 & 43.5 & 49.3 & 48.5 & 50.4 & 41.5 & 58.0 & 35.0 & 28.7 & 24.2 & 32.5 & 36.7 \\
 & TENT~\cite{wang2020tent} & 7.6 & 8.2 & 8.3 & \underline{51.7} & 8.2 & \underline{52.5} & 53.3 & 52.2 & 36.3 & 47.9 & 62.6 & \bf{52.5} & 36.4 & 40.0 & 38.1 & 37.0 \\
 & TPT~\cite{shu2022test} & 17.9 & 19.5 & 27.1 & 43.5 & 20.0 & 42.6 & 48.7 & 49.1 & 49.5 & 42.2 & 57.3 & 33.4 & 31.1 & 27.6 & 32.8 & 36.1 \\
 & WATT~\cite{osowiechi2024watt} & 25.6 & 27.3 & 31.5 & 48.5 & 23.3 & 48.4 & 53.1 & 52.9 & 52.3 & 48.1 & 62.5 & 45.1 & 35.5 & 36.7 & 37.8 & 41.9 \\
 & Mint~\cite{bao2026mint} & 27.5 & 30.5 & 36.0 & 49.3 & 26.5 & 47.4 & 53.4 & 52.6 & 52.1 & \underline{48.5} & \underline{64.7} & 44.1 & 35.2 & 33.0 & 35.4 & 42.4 \\
 & BATCLIP~\cite{maharana2025batclip} & 25.5 & 28.4 & 34.6 & 49.7 & 26.5 & 48.8 & \underline{54.8} & 51.9 & 51.6 & 48.3 & 62.8 & 45.7 & 34.8 & 33.0 & 37.2 & 42.2 \\ 
 & SubTTA~\cite{zeng2026subspace} & \underline{31.9} & \underline{34.2} & \underline{36.3} & 49.5 & \underline{29.7} & 48.9 & 53.9 & \underline{53.6} & \underline{54.1} & 48.2 & 63.2 & 46.5 & \underline{36.5} & \underline{40.2} & \underline{40.8} & \underline{44.5} \\
  & \textbf{CDIM (Ours)} & \bf{33.9} & \bf{36.5} & \bf{42.8} & \bf{53.4} & \bf{32.5} & \bf{52.7} & \bf{58.0} & \bf{56.6} & \bf{56.5}& \bf{53.2} & \bf{67.4} & \underline{52.2} & \bf{41.1} & \bf{44.0} & \bf{41.7} & \bf{48.2} \\
\bottomrule
\end{tabular}
} 
\vspace{-10pt}
\end{table*}
\section{Experiments}
\label{sec:experiment}
 
We first report accuracy under the three standard TTA protocols on three corruption benchmarks (Section~\ref{sec:results}), and then examine the contribution of each component (Section~\ref{sec:empirical}).

\textbf{Dataset and Setup.}
We evaluate \textsc{CDIM} on three standard corruption benchmarks, CIFAR10-C, CIFAR100-C, and ImageNet-C~\cite{hendrycks2019benchmarking}, which are built on CIFAR~\cite{krizhevsky2009learning} and ImageNet~\cite{deng2009imagenet}. Following~\cite{wang2020tent}, we report top-1 accuracy at the highest corruption severity (level 5). The model is reset before each of the 15 corruption types (online setting). We use ViT-B/16 as the default CLIP backbone. 
We use an SSL model that was not fine-tuned on ImageNet or on any other evaluated dataset, so that no label information of the evaluated datasets enters the auxiliary backbone.
Further details are provided in Appendix~\ref{app:detail_set}.

\paragraph{Implementation Details.}
We follow the optimizer setup of~\cite{maharana2025batclip}: AdamW with a learning rate of $10^{-3}$ (CIFAR10-C), Adam with $5\times10^{-4}$ (CIFAR100-C), and AdamW with $5\times10^{-4}$ (ImageNet-C), with batch sizes of 200, 200, and 64, respectively. The hyperparameters of \textsc{CDIM} are fixed across benchmarks: $\tau_D^{-1}=20$ (Eq.~\eqref{eq:logits}), $\rho=0.05$ (Eq.~\eqref{eq:gate}), and $\alpha=0.9$ (Eq.~\eqref{eq:ema}). 
Baselines are run with their original settings, and no target labels are used for any selection (details in Appendix~\ref{app:detail_set}).

\subsection{Results on Test-Time Adaptation}
\label{sec:results}

\paragraph{Online TTA.}
Table~\ref{tab:benchmark_all} reports the online TTA results on the three benchmarks. \textsc{CDIM} improves over the previous state-of-the-art baseline~\cite{zeng2026subspace} by +4.8\% on ImageNet-C, +3.5\% on CIFAR10-C, and +3.7\% on CIFAR100-C. 
The gains are attributable to three components, which are examined in Tables~\ref{tab:ablation} and~\ref{tab:abl_gtim}. First, the cross-modal logit ensemble of Eq.~\eqref{eq:ensemble} yields a better-calibrated posterior than the CLIP+TENT baseline (Figure~\ref{fig:calibration}). Second, the adaptive confidence gate of Section~\ref{sec:tsim} excludes the most uncertain samples from the conditional-entropy gradient. Third, the marginal-entropy term of GT-IM keeps the predicted label distribution from collapsing onto a biased subset of text anchors. With a batch size of 50 to 100, the CIFAR10-C accuracy of CDIM exceeds 81\% (Table~\ref{tab:batch}, Appendix~\ref{sub:bat}). Table~\ref{tab:benchmark_all} nevertheless reports $B = 200$ to follow the protocol of~\cite{maharana2025batclip}. The baselines are not re-tuned for batch size.

\begin{table*}[t]
\centering
\label{tab:continual_mixed}
\vspace{7pt}
\caption{Accuracy (\%) under (a) continual and (b) mixed-domain protocols. Continual: CLIP ViT-B/16. Mixed-domain: backbones follow \cite{bao2026mint}. CDIM uses DINO\,v2-base.}
\begin{subtable}[t]{0.48\linewidth}
\centering
\scriptsize
\setlength{\tabcolsep}{3pt}
\renewcommand{\arraystretch}{0.95}
\begin{tabular}{@{}lccc@{}}
\toprule
Method & CIFAR10-C & CIFAR100-C & ImageNet-C \\
\midrule
TENT~\cite{wang2020tent}  &61.4  &31.7  &23.5  \\
EATA~\cite{niu2022efficient} & 63.5 & 45.8 & 34.1 \\
SAR~\cite{niu2023towards}   & 62.7 & 37.7 & 28.8 \\
COME~\cite{zhang2024come}  & 63.4 & 42.4 & 33.9 \\
BATCLIP~\cite{maharana2025batclip}  & 57.1 & 27.0 & 18.4 \\
Mint~\cite{bao2026mint} & 67.4 & 34.5 & 25.3 \\
\textbf{CDIM (Ours)}  & \textbf{80.4} & \textbf{46.3} & \textbf{34.6} \\
\bottomrule
\end{tabular}
\vspace{1pt}
\caption{Continual}
\label{tab:continual}
\end{subtable}\hfill\begin{subtable}[t]{0.48\linewidth}
\centering
\scriptsize
\setlength{\tabcolsep}{3pt}
\renewcommand{\arraystretch}{0.95}
\begin{tabular}{@{}lccc@{}}
\toprule
Method & CIFAR10-C & CIFAR100-C & ImageNet-C \\
\midrule
CLIP~\cite{radford2021learning}     & 59.0 & 35.8 & 39.6 \\
TDA~\cite{karmanov2024efficient}      & 62.1 & 38.3 & 42.3 \\
DMN-ZS~\cite{zhang2024dual}   & 60.2 & 36.0 & 39.9 \\
WATT-S~\cite{osowiechi2024watt}   & 63.6 & 39.0 & 43.9 \\
CLIPArTT~\cite{a2025clipartt} & 56.9 & 38.7 & 40.5 \\
Mint~\cite{bao2026mint}     & \underline{65.9} & \underline{39.8} & \underline{45.2} \\
\textbf{CDIM (Ours)}     & \bf{71.2} & \bf{40.1}   & \bf{46.9}   \\
\bottomrule
\end{tabular}
\vspace{1pt}
\caption{Mixed-domain}
\label{tab:mixed}
\end{subtable}
\vspace{-7pt}
\end{table*}

\paragraph{Continual TTA.}
In the continual setting~\cite{wang2022continual}, the 15 corruptions are applied sequentially without parameter resets. All methods use CLIP ViT-B/16 (Table~\ref{tab:continual}). \textsc{CDIM} is not designed as a continual TTA method, yet it performs stably in this setting compared with the other approaches. It outperforms the strongest baseline by 13.0\% on CIFAR10-C and by 0.5\% on both CIFAR100-C and ImageNet-C.
This stability may be explained by the frozen DINOv2 anchor and by the online logit adjustment of Section~\ref{sec:ola}, which corrects the residual skew accumulated over long streams.
\paragraph{Mixed-domain.}
Following the mixed-domain setup of~\cite{niu2023towards,bao2026mint}, we mix all 15 corruption types into a single evaluation stream and use three different CLIP backbones, as in~\cite{bao2026mint}. As shown in Table~\ref{tab:mixed}, \textsc{CDIM} outperforms Mint~\cite{bao2026mint} by +5.3\% on CIFAR10-C (ViT-B/32), +0.3\% on CIFAR100-C (ViT-B/16), and +1.7\% on ImageNet-C (ViT-L/14). 
The margin is largest on CIFAR10-C ($K = 10$), which we attribute to the marginal entropy $H(\bar{\mathbf{p}})$ penalizing prediction skew more strongly when $K$ is small. 
For larger $K$, the online logit adjustment may play the main stabilizing role by correcting the skew accumulated over the mixed stream.
\begin{table}[t!]
\centering
\scriptsize
\vspace{-18pt}
\caption{Accuracy (\%) under various settings. Default: CLIP ViT-B/16 with DINO\,v2-base. (a) Auxiliary self-supervised model and (b) DINO\,v2 scale for each CLIP backbone, on CIFAR10-C. (c) TinyCLIP (ViT-61M/32) on CIFAR10-C. (d) ImageNet variants with CLIP ViT-B/16.}
\label{tab:variation}
\vspace{5pt}
\begin{subtable}[t]{0.47\linewidth}
\centering
\scriptsize
\setlength{\tabcolsep}{3pt}
\renewcommand{\arraystretch}{0.9}
\begin{tabular}{@{}cccc@{}}
\toprule
\multicolumn{4}{c}{\bf{(a) Self-supervised model}} \\
\midrule
DINO\,v1 & DINO\,v3 & MoCo\,v3 & MAE \\
\midrule
78.8 & 79.7 & 79.1 & 79.3 \\
\specialrule{0.08em}{0.4em}{0pt}
\specialrule{0.08em}{0.4em}{0.4em}
\multicolumn{4}{c}{\bf{(b) Model scale}} \\
\midrule
DINO\,v2 & small & base & large \\
\midrule
ViT-B/32 & 76.6 & 77.1 & 77.4 \\
ViT-B/16 & 79.1 & 79.8 & 80.3 \\
ViT-L/14 & 87.1 & 87.7 & 88.7 \\
\bottomrule
\end{tabular}
\vspace{-3pt}
\label{tab:multi}
\end{subtable}%
\hfill%
\begin{subtable}[t]{0.51\linewidth}
\centering
\scriptsize
\setlength{\tabcolsep}{3pt}
\renewcommand{\arraystretch}{0.9}
\begin{tabular}{@{}lcccc@{}}
\toprule
\multicolumn{5}{c}{\bf{(c) TinyCLIP (ViT-61M/32)}} \\
\midrule
Model & CLIP & TENT & BATCLIP & \textbf{CDIM} \\
\midrule
ViT-61M/32 & 63.9 & 65.7 & 72.8 & 78.8 \\
\specialrule{0.08em}{0.4em}{0pt}
\specialrule{0.08em}{0.4em}{0.4em}
\multicolumn{5}{c}{\bf{(d) ImageNet variants (ViT-B/16)}} \\
\midrule
Method & IN-A & IN-V2 & IN-R & IN-S \\
\midrule
Mint~\cite{bao2026mint} & 54.7 & 62.6 & 78.1 & 48.4 \\
BATCLIP~\cite{maharana2025batclip} & 48.7 & 60.5 & 72.6 & 39.9 \\
\textbf{CDIM (Ours)} & \textbf{55.5} & \textbf{63.3} & \textbf{79.2} & \textbf{50.1} \\
\bottomrule
\end{tabular}
\vspace{-3pt}

\label{tab:amulti}
\end{subtable}
\vspace{-10pt}
\end{table}

\subsection{Empirical Studies}
\label{sec:empirical}

\begin{wraptable}{r}{0.45\textwidth}
\centering
\scriptsize
\vspace{-13pt}
\caption{Ablation study on the contributions of CDIM. The first row is the CLIP+TENT.}
\vspace{-4pt}
\label{tab:ablation}
\setlength{\tabcolsep}{2pt}
\renewcommand{\arraystretch}{0.95}
\begin{tabular}{@{}ccc ccc@{}}
\toprule
ENS & GT-IM & OLA & ImageNet-C & CIFAR10-C & CIFAR100-C \\
\midrule
 &  &  & 25.2 & 61.9 & 37.0 \\
\checkmark &  & &28.7 & 63.1 & 40.0 \\
 & \checkmark &  &33.8 & 76.9 & 45.2 \\
 & \checkmark & \checkmark & 35.0 & 78.3 & 46.2 \\
\checkmark & \checkmark &  & 36.3 & 79.2 & 47.5 \\
\checkmark & \checkmark & \checkmark & \textbf{36.9} & \textbf{79.8} & \textbf{48.2} \\
\bottomrule
\end{tabular}
    \vspace{-1\baselineskip}
\end{wraptable}

\paragraph{Ablation of each component.}
Table~\ref{tab:ablation} evaluates the components of \textsc{CDIM} with CLIP+TENT as the baseline. Adding the DINO\,v2 ensemble alone gives a consistent improvement, whereas GT-IM yields the largest single gain (e.g., +15.0\% on CIFAR10-C). Combining GT-IM with either the cross-modal ensemble or the online logit adjustment (OLA) further raises accuracy, and the full \textsc{CDIM} achieves the highest accuracy on all three benchmarks. These results indicate that the three components contribute complementary gains. Appendix~\ref{app:ablations} decomposes GT-IM and compares \textsc{CDIM} with baselines that use the same DINO\,v2 ensemble.

\paragraph{Variation of experiments.}
Table~\ref{tab:variation}(a) examines the dependence of \textsc{CDIM} on the auxiliary self-supervised model. 
Replacing DINO\,v2 with DINO\,v1, DINO\,v3, MoCo\,v3, or MAE changes the CIFAR10-C accuracy by at most 1.0\%, which indicates that the improvement does not rely on one auxiliary model.
 In Table~\ref{tab:variation}(b), the accuracy increases monotonically with the scale of the CLIP encoder (by 10.6\% from ViT-B/32 to ViT-L/14 with DINO\,v2-base) and with the scale of the auxiliary DINO\,v2 (by up to 1.6\% from small to large). \textsc{CDIM} benefits from larger backbones without accuracy degradation.
  On ImageNet (IN) variations, \textsc{CDIM} improves over Mint by up to 1.7\% and over BATCLIP by up to 10.2\% (Table~\ref{tab:variation}(d)), 
 which indicates that the improvement extends beyond synthetic corruptions to natural distribution shifts, including natural adversarial examples.
 With the lightweight TinyCLIP ViT-61M/32 encoder~\cite{wu2023tinyclip} on CIFAR10-C, \textsc{CDIM} exceeds the BATCLIP baseline by 6.0\% (Table~\ref{tab:variation}(c)), which indicates that the gain is retained with a smaller encoder.

\begin{figure*}[t]
\centering
\begin{subtable}[t]{0.33\linewidth}
  \centering
  \includegraphics[width=\linewidth,height=3.55cm]{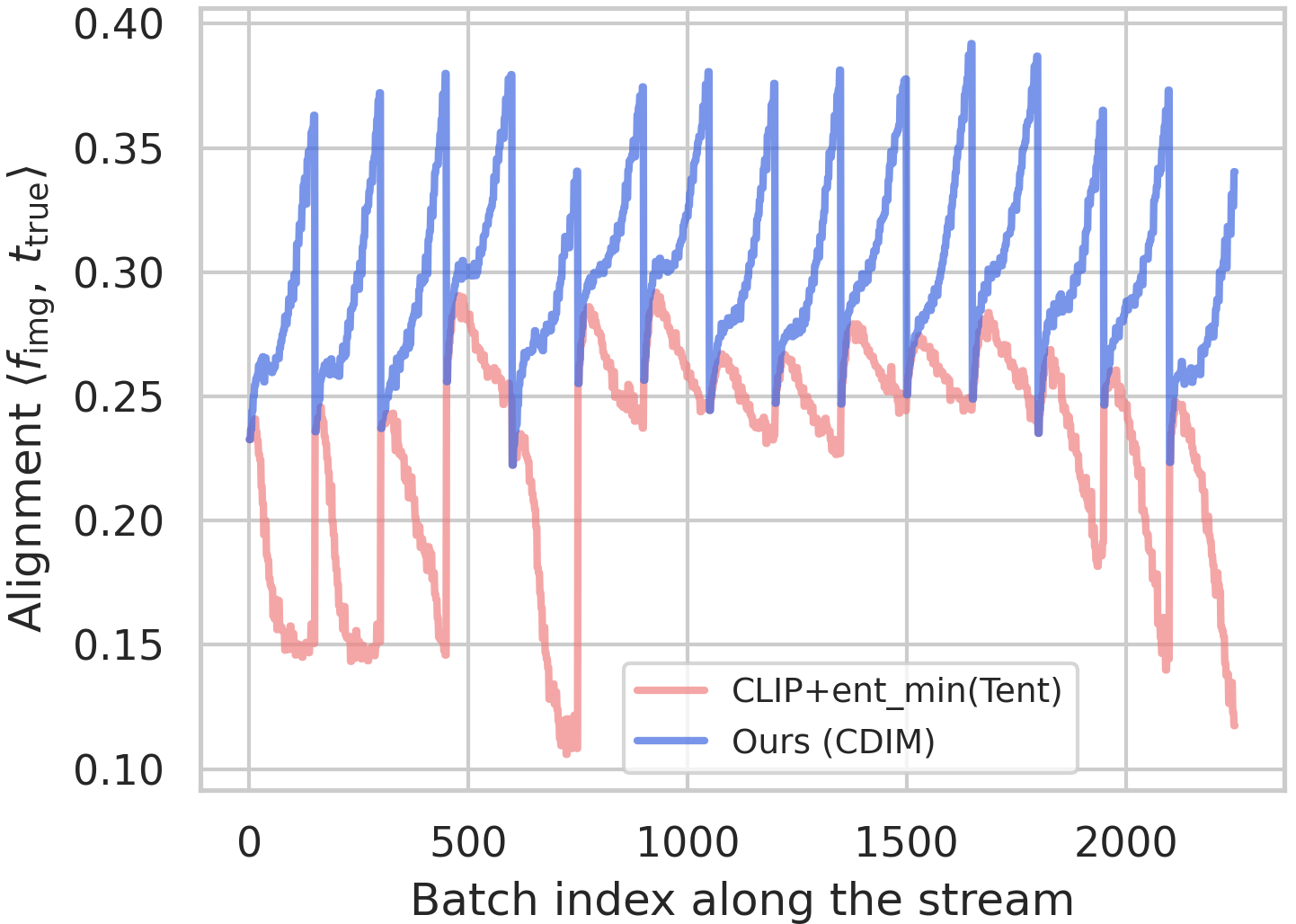}
  \caption{Image-Text Alignment over time}
  \label{fig:separability}\label{fig:separability-left}
\end{subtable}\hfill
\begin{subtable}[t]{0.32\linewidth}
  \centering
  \includegraphics[width=\linewidth]{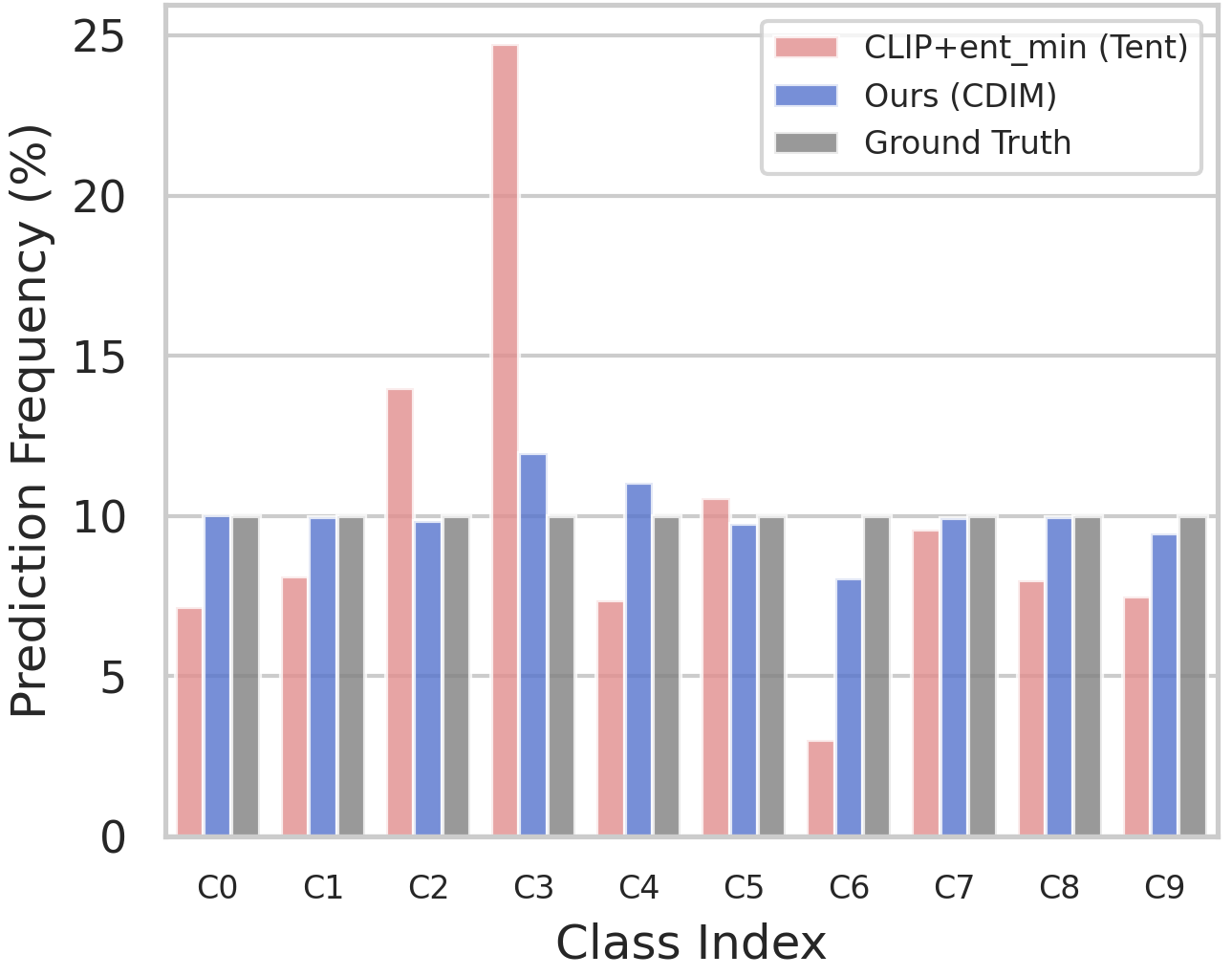}
  \caption{Prediction Skewness}
  \label{fig:skew}
\end{subtable}\hfill
\begin{subtable}[t]{0.32\linewidth}
  \centering
  \includegraphics[width=\linewidth,height=3.55cm]{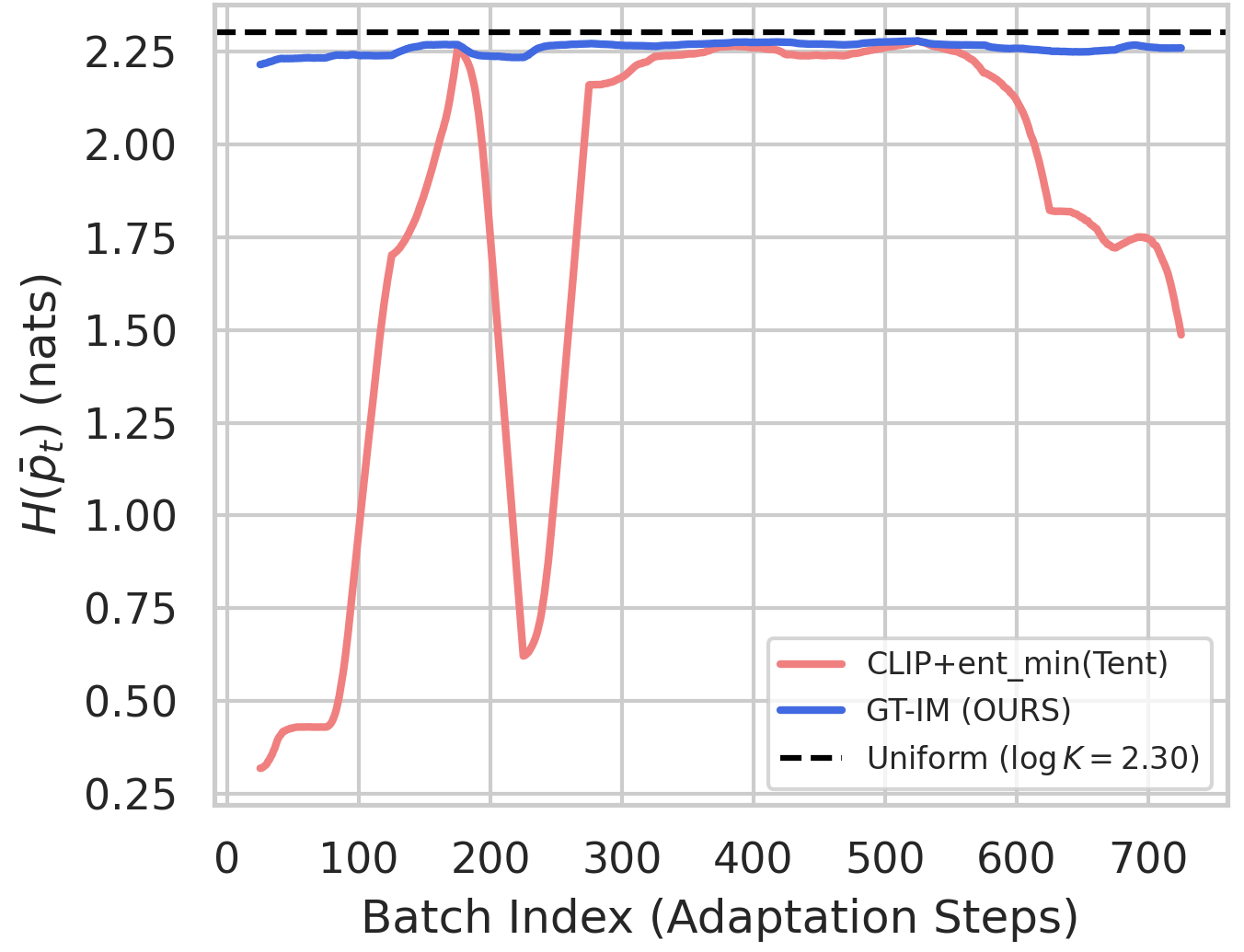}
  \caption{Marginal Entropy Drift}
  \label{fig:drift}
\end{subtable}
\caption{Three analyses during the CIFAR10-C TTA process, comparing CLIP+TENT and \textsc{CDIM}. \textbf{(a)} Image-text alignment of CLIP, measured as $\langle\mathbf{f}_{\text{img}}, \mathbf{t}_{\text{true}}\rangle$ per batch during TTA.
\textbf{(b)} Cumulative prediction frequency per class, with the ground-truth frequency for reference. \textbf{(c)} Marginal entropy $H(\bar{\mathbf{p}}_t)$ per batch; the dashed line marks $\log K$.}
    \vspace{-10pt}
\end{figure*}

\textbf{Changes in cross-modal alignment over TTA training.}
Figure~\ref{fig:separability} tracks the cosine alignment $\langle f_{\mathrm{img}}, t_{\mathrm{true}}\rangle$, which measures how well the adapted image features ($f_{\mathrm{img}}$) align with the correct text anchor ($t_{\mathrm{true}}$) during adaptation on CIFAR10-C. Under CLIP+TENT~\cite{wang2020tent}, the alignment drifts downward and recovers slowly after each corruption transition, which is the empirical counterpart of the image-text alignment erosion reported in~\cite{maharana2025batclip,zeng2026subspace}. Under \textsc{CDIM}, the alignment is higher throughout and recovers within a small number of batches, which is consistent with the stabilizing effect of the frozen DINO\,v2 anchor.

\begin{wrapfigure}{r}{0.3\textwidth}
    \vspace{-1\baselineskip}
    \centering
    \includegraphics[width=0.44\textwidth,height=6.3cm]{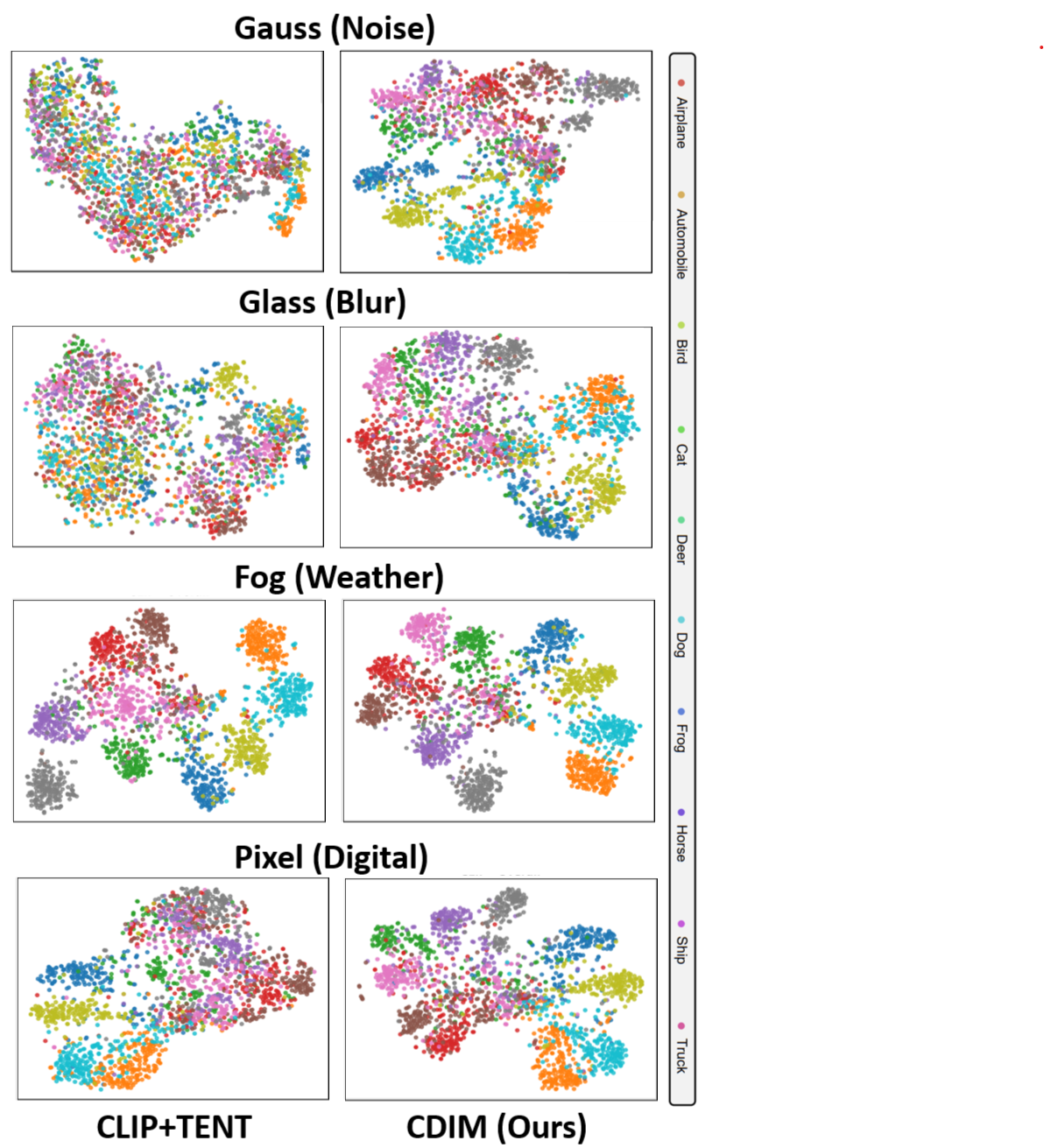}
    \vspace{-1.5\baselineskip}
    \caption{Comparison t-SNE visualization (CIFAR10-C).}
    \vspace{-2\baselineskip}
    \label{fig:tsne}
\end{wrapfigure}

 \paragraph{Detailed Analysis of \textsc{CDIM}.}
We analyze the adaptation dynamics of \textsc{CDIM} on CIFAR10-C. Figure~\ref{fig:skew} shows that CLIP+TENT suffers from a severe class-distribution skew, with about 25\% of its predictions falling on class C3, whereas \textsc{CDIM} maintains a near-uniform distribution. This is consistent with OLA and GT-IM jointly preventing label-set collapse. 
The marginal entropy $H(\bar p_t)$ in Figure~\ref{fig:drift} supports this observation: it drops sharply and drifts under CLIP with entropy minimization, but stays close to the uniform reference ($\log K = 2.30$) under \textsc{CDIM}. The t-SNE projections of one representative case per corruption category (Figure~\ref{fig:tsne}) show a similar picture. Under severe corruptions, the baseline embeddings collapse into a mixed cluster, whereas CDIM preserves clear class separation. These analyses compare the full \textsc{CDIM} with CLIP+TENT, and Table~\ref{tab:ablation} reports the effect of each component.

\paragraph{Additional experiments.}
We further evaluate \textsc{CDIM} on domain generalization benchmarks~\cite{gulrajani2020search} (Appendix~\ref{add_domain}) and on ImageNet variants with natural distribution shifts and natural adversarial examples (Appendix~\ref{sec:appendix_imagenet_variants}). Detailed ablation studies and analyses are provided in Appendices~\ref{app:ablations} and~\ref{sec:new-ablation}, and the computational overhead, including GPU memory and latency, is analyzed in Appendix~\ref{app:Resource}.
\section{Conclusion}
\label{sec:conclusion}
We present \textsc{CDIM}, an online TTA method that mitigates modality bias, a contributor to confirmation bias in CLIP-based TTA, at three complementary levels. 
First, a cross-modal logit ensemble with a frozen DINO\,v2 backbone stabilizes the representation. Second, a gated information-maximization objective prevents label collapse and excludes unreliable samples from the gradient updates. 
Third, an online logit adjustment corrects the residual prediction skew. 
When all methods are evaluated with CLIP, \textsc{CDIM} outperforms the compared baselines on the three corruption benchmarks in the online TTA setting and achieves competitive performance in the continual and mixed-domain settings. These gains require one additional frozen DINOv2 forward pass per batch and gradient-free corrections.

\section*{Acknowledgments}
This work was supported by the IITP grant funded by the Korea government (MSIT) (RS-2025-02283048, Developing the Next-Generation General AI with Reliability, Ethics, and Adaptability and RS-2023-00236245, Development of Perception/Planning AI SW for Seamless Autonomous Driving in Adverse Weather/Unstructured Environment), National Research Foundation of Korea (NRF) (RS-2026-25475721 and RS-2025-22712971) and the Technology Innovation Program (RS-2025-02220839) funded by the Ministry of Trade, Industry and Energy (MOTIE, Korea).

\bibliographystyle{plain} 
\bibliography{neurips_2026.bib}

%%%%%%%%%%%%%%%%%%%%%%%%%%%%%%%%%%%%%%%%%%%%%%%%%%%%%%%%%%%%

%\appendix

% \section{Technical appendices and supplementary material}
% Technical appendices with additional results, figures, graphs, and proofs may be submitted with the paper submission before the full submission deadline (see above). You can upload a ZIP file for videos or code, but do not upload a separate PDF file for the appendix. There is no page limit for the technical appendices. 

% Note: Think of the appendix as ``optional reading'' for reviewers. The paper must be able to stand alone without the appendix; for example, adding critical experiments that support the main claims to an appendix is inappropriate. 

%%%%%%%%%%%%%%%%%%%%%%%%%%%%%%%%%%%%%%%%%%%%%%%%%%%%%%%%%%%%

\newpage
\appendix
\section*{NeurIPS Paper Checklist}

\begin{enumerate}

\item {\bf Claims}
    \item[] Question: Do the main claims made in the abstract and introduction accurately reflect the paper's contributions and scope?
    \item[] Answer: \answerYes{}%\answerTODO{} % Replace by \answerYes{}, \answerNo{}, or \answerNA{}.
    \item[] Justification: In the abstract and introduction, we claim that this paper focuses on test-time
adaptation of vision-language models. Contributions are clearly listed at the end of the introduction, at the beginning of the proposed method and in the abstract.

\item {\bf Limitations}
    \item[] Question: Does the paper discuss the limitations of the work performed by the authors?
    \item[] Answer: \answerYes{}%\answerTODO{} % Replace by \answerYes{}, \answerNo{}, or \answerNA{}.
    \item[] Justification: Please refer to Appendix~\ref{limit}. %\justificationTODO{}

\item {\bf Theory assumptions and proofs}
    \item[] Question: For each theoretical result, does the paper provide the full set of assumptions and a complete (and correct) proof?
    \item[] Answer: \answerNA{}%\answerTODO{} % Replace by \answerYes{}, \answerNo{}, or \answerNA{}.
    \item[] Justification:  The paper does not include formal theoretical results such as theorems or proofs. The assumptions underlying the method, namely a uniform target class prior and a source prior estimated from the predictions of the model itself, are described in Sections~\ref{sec:intro} and~\ref{sec:method} (Eqs.~\eqref{eq:prior_shift} to~\eqref{eq:ema}), and their scope is discussed in Appendix~\ref{limit}. The effect of the resulting correction is examined empirically in Section~\ref{sec:experiment} and Appendix~\ref{app:ablations}.

    \item {\bf Experimental result reproducibility}
    \item[] Question: Does the paper fully disclose all the information needed to reproduce the main experimental results of the paper to the extent that it affects the main claims and/or conclusions of the paper (regardless of whether the code and data are provided or not)?
    \item[] Answer: \answerYes{}%\answerTODO{} % Replace by \answerYes{}, \answerNo{}, or \answerNA{}.
    \item[] Justification: Section~\ref{sec:experiment} and Appendix~\ref{app:detail_set} specify the backbones, the optimizers, learning rates and batch sizes, the hyperparameters, and the three evaluation protocols. Algorithm~\ref{alg:cdim} (Appendix~\ref{app:pseudo}) gives pseudo-code for one update.

\item {\bf Open access to data and code}
    \item[] Question: Does the paper provide open access to the data and code, with sufficient instructions to faithfully reproduce the main experimental results, as described in supplemental material?
    \item[] Answer: \answerYes{} %\answerYes{}%\answerTODO{} % Replace by \answerYes{}, \answerNo{}, or \answerNA{}.
    \item[] Justification: We use public datasets (CIFAR10-C, CIFAR100-C, ImageNet-C, ImageNet variants, VisDA-C, OfficeHome, PACS, and VLCS) and provide pseudocode (Algorithm~\ref{alg:cdim} in Appendix~\ref{app:pseudo}) and a GitHub repository.

\item {\bf Experimental setting/details}
    \item[] Question: Does the paper specify all the training and test details (e.g., data splits, hyperparameters, how they were chosen, type of optimizer) necessary to understand the results?
    \item[] Answer: \answerYes{} %\answerTODO{} % Replace by \answerYes{}, \answerNo{}, or \answerNA{}.
    \item[] Justification: Section~\ref{sec:experiment} and Appendix~\ref{app:detail_set} specify the datasets, corruption severity, optimizers, learning rates, batch sizes, and the hyperparameters of CDIM together with how they were set. Appendices~\ref{subsec:abl-ema-momentum} and~\ref{subsec:abl-discard-ratio} report post-hoc sensitivity sweeps that were not used for selection.

\item {\bf Experiment statistical significance}
    \item[] Question: Does the paper report error bars suitably and correctly defined or other appropriate information about the statistical significance of the experiments?
    \item[] Answer: \answerYes{}%\answerTODO{} % Replace by \answerYes{}, \answerNo{}, or \answerNA{}.
    \item[] Justification: The main results in Tables~\ref{tab:benchmark_all},~\ref{tab:continual} and~\ref{tab:mixed} use a single fixed seed, following prior work. For CDIM, Appendix~\ref{sub:bat} (Table~\ref{tab:batch}) reports the mean and standard deviation over five random seeds for each corruption type and batch size. At the default batch size, the five-seed means match the fixed-seed results within 0.2 points. We also include a statistical significance analysis in Section~\ref{sec:experiment} and in Appendix~\ref{sec:new-ablation}.


\item {\bf Experiments compute resources}
    \item[] Question: For each experiment, does the paper provide sufficient information on the computer resources (type of compute workers, memory, time of execution) needed to reproduce the experiments?
    \item[] Answer: \answerYes{} %\answerTODO{} % Replace by \answerYes{}, \answerNo{}, or \answerNA{}.
    \item[] Justification: All experiments were run on a single NVIDIA V100 GPU. Appendix~\ref{app:Resource} and Table~\ref{tab:eff} report per-batch latency, peak GPU memory, and throughput for CDIM, TENT, and BATCLIP on CIFAR100-C.%\justificationTODO{}
    
\item {\bf Code of ethics}
    \item[] Question: Does the research conducted in the paper conform, in every respect, with the NeurIPS Code of Ethics \url{https://neurips.cc/public/EthicsGuidelines}?
    \item[] Answer: \answerYes{} %\answerTODO{} % Replace by \answerYes{}, \answerNo{}, or \answerNA{}.
    \item[] Justification: We strictly adhere to the NeurIPS Code of Ethics.

\item {\bf Broader impacts}
    \item[] Question: Does the paper discuss both potential positive societal impacts and negative societal impacts of the work performed?
    \item[] Answer: \answerYes{} %\answerTODO{} % Replace by \answerYes{}, \answerNo{}, or \answerNA{}.
    \item[] Justification: Please refer to Appendix~\ref{app:impact}. %\justificationTODO{}
    
\item {\bf Safeguards}
    \item[] Question: Does the paper describe safeguards that have been put in place for responsible release of data or models that have a high risk for misuse (e.g., pre-trained language models, image generators, or scraped datasets)?
    \item[] Answer: \answerNA{} %\answerTODO{} % Replace by \answerYes{}, \answerNo{}, or \answerNA{}.
    \item[] Justification: Our paper does not release new data or models.


\item {\bf Licenses for existing assets}
    \item[] Question: Are the creators or original owners of assets (e.g., code, data, models), used in the paper, properly credited and are the license and terms of use explicitly mentioned and properly respected?
    \item[] Answer: \answerYes{} %\answerTODO{} % Replace by \answerYes{}, \answerNo{}, or \answerNA{}.
    \item[] Justification: All assets used in this work are properly credited, and their respective licenses and terms of use are  observed.

\item {\bf New assets}
    \item[] Question: Are new assets introduced in the paper well documented and is the documentation provided alongside the assets?
    \item[] Answer: \answerNA{} %\answerTODO{} % Replace by \answerYes{}, \answerNo{}, or \answerNA{}.
    \item[] Justification: The paper releases no new datasets or pretrained models. 

\item {\bf Crowdsourcing and research with human subjects}
    \item[] Question: For crowdsourcing experiments and research with human subjects, does the paper include the full text of instructions given to participants and screenshots, if applicable, as well as details about compensation (if any)? 
    \item[] Answer: \answerNA{} %\answerTODO{} % Replace by \answerYes{}, \answerNo{}, or \answerNA{}.
    \item[] Justification: Our paper does not involve crowdsourcing nor research with human subjects.

\item {\bf Institutional review board (IRB) approvals or equivalent for research with human subjects}
    \item[] Question: Does the paper describe potential risks incurred by study participants, whether such risks were disclosed to the subjects, and whether Institutional Review Board (IRB) approvals (or an equivalent approval/review based on the requirements of your country or institution) were obtained?
    \item[] Answer: \answerNA{} %\answerTODO{} % Replace by \answerYes{}, \answerNo{}, or \answerNA{}.
    \item[] Justification: Our paper does not involve crowdsourcing nor research with human subjects.

\item {\bf Declaration of LLM usage}
    \item[] Question: Does the paper describe the usage of LLMs if it is an important, original, or non-standard component of the core methods in this research? Note that if the LLM is used only for writing, editing, or formatting purposes and does \emph{not} impact the core methodology, scientific rigor, or originality of the research, declaration is not required.
    \item[] Answer: \answerNA{} %\answerTODO{} % Replace by \answerYes{}, \answerNo{}, or \answerNA{}.
    \item[] Justification: LLMs were used only for writing, editing, and formatting, and were not used for the core methodology, scientific rigor, or originality of the research.

\end{enumerate}
\newpage
\noindent\textbf{\Large Appendix}

\appendix

\section{Additional Discussion}

\subsection{Limitations}
\label{limit}
The discard ratio $\rho$ is fixed across benchmarks,
so the quantile gate always discards $\lfloor \rho B \rfloor$ samples per batch (at least one), regardless of how many unreliable samples the batch contains. 
This can remove correct samples under mild corruption, and it may also exclude samples of unseen classes when the stream contains out-of-distribution classes.
A severity- or dispersion-adaptive $\rho$ is left for future work. 
The DINO-branch inverse temperature $\tau_D^{-1}$ is likewise a fixed constant and was not tuned. Like $\rho$ and $\alpha$, it follows common conventions~\cite{caron2021emerging,oquab2023dinov2,shu2022test, kingma2014adam} rather than a per-benchmark search, while the optimizer, learning rate, and batch size follow BATCLIP~\cite{maharana2025batclip} (Appendix~\ref{app:detail_set}).
A second limitation concerns the class distribution. Both the marginal term of GT-IM and OLA assume an approximately uniform target class distribution. Accuracy may therefore degrade on class-imbalanced or temporally correlated streams, and a more robust class-prior estimator would be required.
Beyond accuracy, the auxiliary DINO\,v2 branch adds latency and memory overhead compared with single-backbone methods such as TENT (Appendix~\ref{app:Resource}). 
Extending the framework beyond image classification, for example to object detection and semantic segmentation, is left for future work.

\subsection{Broader Impacts}
\label{app:impact}
Our work addresses modality bias and the resulting confirmation bias in vision-language models under distribution shift. Improving robustness under test-time distribution shift can support the reliability of applications such as autonomous driving, robotics, and video analysis, where the data distribution changes over time. \textsc{CDIM} relies on a frozen self-supervised backbone and training-free corrections, but it adds one DINO\,v2 forward pass per batch (Appendix~\ref{app:Resource}). Therefore, \textsc{CDIM} contributes to safer and more dependable AI deployments in the wild, and with a lightweight encoder such as TinyCLIP, it requires 0.41 times the peak GPU memory of TENT with ViT-B/16 (Appendix~\ref{app:Resource}).

We do not anticipate direct negative societal impacts from the algorithm itself. \textsc{CDIM} operates without labels at test time and does not require access to private user data. Because it builds on large-scale foundation models such as CLIP and DINO\,v2, it inherits the representational and social biases present in their pre-training data. Improved zero-shot robustness could also be misused in surveillance or profiling applications. Further study of the biases of foundation models is needed as adaptation techniques are applied more widely.

\subsection{Detailed Experiments Settings}
\label{app:detail_set}
For a fair comparison with recent state-of-the-art methods such as BATCLIP~\cite{maharana2025batclip} and SubTTA~\cite{zeng2026subspace}, CLIP with a ViT-B/16 backbone is used as the default architecture. For the auxiliary self-supervised learning (SSL) model, DINO\,v2-base (ViT-B/14) is used. The version \textbf{without ImageNet fine-tuning} is used so that only the self-supervised representation is evaluated.

We evaluate our approach on standard benchmarks~\cite{hendrycks2019benchmarking} containing 15 distinct corruption types: Gaussian noise, shot noise, impulse noise, defocus blur, glass blur, motion blur, zoom blur, snow, frost, fog, brightness, contrast, elastic transform, pixelation, and JPEG compression. Unless otherwise stated, all experiments are conducted at the highest corruption severity (level 5). Results for severity levels 1 to 5 are reported in Table~\ref{tab:severity_results} (Appendix~\ref{subsec:new-abl-severity}). We consider three adaptation protocols.

In the standard online TTA protocol~\cite{wang2020tent}, the model parameters and optimizer state are reset between independent corruption types but are preserved across batches within the same corruption. Under the continual TTA (CTTA) protocol~\cite{wang2022continual}, the model is sequentially adapted to the 15 target domains in a continuous stream. The model parameters and the optimizer state are preserved across all batches, regardless of the corruption type. In the mixed-domain setting~\cite{niu2023towards}, the data from all 15 corruptions are shuffled into a single interleaved stream, and adaptation and evaluation are performed on this stream without domain boundaries.
\paragraph{Hyperparameters.}
\textsc{CDIM} uses the same hyperparameters on all benchmarks: an inverse temperature $\tau_D^{-1}=20$ for the DINO branch, a discard ratio of $\rho=0.05$, and an EMA momentum of $\alpha=0.9$. The CLIP branch keeps its pretrained logit scale $\tau_C^{-1}=100$, and neither scale is tuned. These values were set from common conventions before any benchmark evaluation. 
The inverse temperature of 20 corresponds to a temperature of 0.05, which lies within the teacher-temperature range of 0.04 to 0.07 used by DINO~\cite{caron2021emerging} and DINO\,v2~\cite{oquab2023dinov2}. 
The discard ratio $\rho$ is a small fixed fraction, as in the quantile-based confidence selection of~\cite{shu2022test}, and $\alpha = 0.9$ equals the default first-moment decay rate of Adam~\cite{kingma2014adam}. 
The sweeps in Table~\ref{tab:sens_alpha_rho} (Appendices~\ref{subsec:abl-ema-momentum} and~\ref{subsec:abl-discard-ratio}) are post-hoc sensitivity analyses and were not used for selection. 
The optimizer, the learning rate, and the batch size follow BATCLIP~\cite{maharana2025batclip}. 
BATCLIP~\cite{maharana2025batclip} and Mint~\cite{bao2026mint} are CLIP-specific methods, whereas the remaining baselines (TENT~\cite{wang2020tent}, CoTTA~\cite{wang2022continual}, EATA~\cite{niu2022efficient}, SAR~\cite{niu2023towards}, and COME~\cite{zhang2024come}) are general TTA methods that we re-run with CLIP ViT-B/16. The baselines use the default hyperparameters of their authors where available, and no target labels are used for any selection.

 \subsection{Resource and Computational cost}
\label{app:Resource}
All experiments are conducted on a single NVIDIA V100 GPU. Table~\ref{tab:eff} compares the per-batch latency, peak GPU memory, and throughput of \textsc{CDIM} with those of TENT and BATCLIP on CIFAR100-C. ViT-B/16 is used with a batch size of 200, following the protocol of~\cite{bao2026mint}. \textsc{CDIM} with DINO\,v2-small adds one frozen DINO\,v2 forward pass and the cross-covariance projection, which increases the latency by 654 ms per batch (117.8\% over TENT). The peak GPU memory increases by 2248 MB over TENT because of the additional DINO\,v2 model, while the logit adjustment step adds negligible overhead ($<$0.1 ms). With the default DINO\,v2-base, the latency increases by 792 ms (142.7\%) and the peak memory by 3497 MB over TENT. When the lighter TinyCLIP ViT-61M/32~\cite{wu2023tinyclip} is combined with DINO\,v2-small, \textsc{CDIM} reduces the peak GPU memory to 5335 MB, which is 0.41 times that of TENT with ViT-B/16, while the latency is 2.05 times that of TENT. This configuration reaches 45.5\% accuracy, which is 3.3\% higher than BATCLIP (42.2\%) and 8.5\% higher than TENT (37.0\%), both with ViT-B/16.

\begin{table}[t]
\footnotesize
\caption{Efficiency comparison on CIFAR100-C measured on an NVIDIA V100 GPU. }
    \vspace{5pt}
    \centering
    \resizebox{\linewidth}{!}{
    \begin{tabular}{llcccc}
        \toprule
        \textbf{CLIP encoder} & \textbf{SSL model} & \textbf{Accuracy (\%)} & \textbf{Peak mem (MB)} & \textbf{Latency (ms/batch)} & \textbf{Throughput (img/s)} \\
        \midrule
        ViT-B/16 & None (TENT) & 37.0 & 13146 & 555 & 360.6 \\

        ViT-B/16 & None (BATCLIP) & 42.2 & 14505 & 665 & 300.7 \\

        ViT-B/16 & DINO\,v2-base & 48.2 & 16643 & 1347 & 148.5 \\

        ViT-B/16 & DINO\,v2-small & 47.9 & 15394 & 1209 & 165.5 \\

        TinyCLIP ViT-61M/32 & DINO\,v2-base & 46.7 & 5837 & 1277 & 156.7 \\

        TinyCLIP ViT-61M/32 & DINO\,v2-small & 45.5 & 5335 & 1139 & 175.6 \\
        \bottomrule
    \end{tabular}}
    \label{tab:eff}
\end{table}

\section{Additional Ablation Studies}
\label{app:ablations}

\begin{table}[t]
\centering
\scriptsize
\caption{Ablation study of OLA. The first row is the CLIP+TENT baseline.}
\vspace{5pt}
\label{tab:abl_ola}
\setlength{\tabcolsep}{5pt}
\renewcommand{\arraystretch}{0.95}
\begin{tabular}{@{}cccc ccc@{}}
\toprule
ENS (w/ DINO)  &GT-IM & OLA   &ImageNet-C & CIFAR10-C & CIFAR100-C \\
\midrule
 &  &   & 25.2 & 61.9 & 37.0 \\
     &   &\checkmark &30.4  &64.0  &41.2  \\ \hline
 \checkmark &    & &28.7 & 63.1 & 40.0  \\
   \checkmark  &   &\checkmark &33.1  & 66.5 &44.5 \\ \hline
   &  \checkmark & &33.8 & 76.9 & 45.2 \\ 
 &\checkmark  &\checkmark &35.0 & 78.3 & 46.2 \\  \hline 
  \checkmark& \checkmark & &36.3 & 79.2 & 47.5  \\
\checkmark &\checkmark  &\checkmark &\textbf{36.9} & \textbf{79.8} & \textbf{48.2} \\
\bottomrule
\end{tabular}
\end{table}

\subsection{Online Logit Adjustment (OLA) ablation}
Table~\ref{tab:abl_ola} isolates the contribution of the online logit adjustment (OLA) described in Section~\ref{sec:ola}. Removing OLA from the full \textsc{CDIM} pipeline, that is, returning $\arg\max_c Z^{\mathrm{ens}}_{ic}$ instead of applying the adjustment of Eq.~\eqref{eq:ola}, lowers the accuracy by $0.6\%$ on ImageNet-C, $0.6\%$ on CIFAR10-C, and $0.7\%$ on CIFAR100-C. This suggests that a measurable class-distribution skew remains after GT-IM. Applying OLA to the CLIP+TENT baseline, which suffers from stronger modality bias, yields larger gains of $+5.2\%$, $+2.1\%$, and $+4.2\%$ on the three benchmarks. The reliability diagram in Figure~\ref{fig:app_abl2_real} points in the same direction: adding OLA to CLIP+TENT lowers the Expected Calibration Error (ECE) from $0.400$ to $0.281$, although some overconfidence on false positives remains. These results are consistent with OLA reducing the prediction skew induced by modality bias.

\renewcommand{\meansd}[2]{#1{\scriptsize{ (#2)}}}

\begin{table}[t!]
    \centering
    \caption{Accuracy (mean (std) \%) on OTTA corruption benchmarks with different batch sizes in ViT-B/16. }
        \vspace{5pt}
    \label{tab:batch}
    \resizebox{1.0\linewidth}{!}{
    \setlength{\tabcolsep}{1.5mm}{
        \begin{tabular}{ccccccccccccccccc}
            \toprule
            & \multicolumn{16}{c}{CIFAR10-C (fixed seed: 79.8)} \\
            \cmidrule(lr){2-17}
            Batch Size & \multicolumn{3}{c}{Noise} & \multicolumn{4}{c}{Blur} & \multicolumn{4}{c}{Weather} & \multicolumn{4}{c}{Digital} & \multirow{2.5}{*}{Avg.} \\
            \cmidrule(lr){2-4} \cmidrule(lr){5-8} \cmidrule(lr){9-12} \cmidrule(lr){13-16}
            & Gauss. & Shot & Impul. & Defoc. & Glass & Motion & Zoom & Snow & Frost & Fog & Brit. & Contr. & Elastic & Pixel & JPEG & \\
            \midrule
            20 & \meansd{67.3}{0.6} & \meansd{70.2}{0.6} & \meansd{75.5}{0.8} & \meansd{82.5}{0.4} & \meansd{68.2}{0.7} & \meansd{82.7}{0.3} & \meansd{84.6}{0.7} & \meansd{85.0}{0.7} & \meansd{85.1}{0.6} & \meansd{85.5}{0.8} & \meansd{91.5}{0.3} & \meansd{88.1}{0.5} & \meansd{74.4}{0.5} & \meansd{79.8}{0.4} & \meansd{72.1}{1.2} & \meansd{79.5}{0.2} \\
            50 & \meansd{69.5}{0.6} & \meansd{71.9}{0.8} & \meansd{78.1}{0.6} & \meansd{84.7}{0.5} & \meansd{70.0}{0.6} & \meansd{84.2}{0.2} & \meansd{86.3}{0.1} & \meansd{87.1}{0.2} & \meansd{87.3}{0.3} & \meansd{86.7}{0.4} & \meansd{93.3}{0.2} & \meansd{90.0}{0.2} & \meansd{76.9}{0.5} & \meansd{81.0}{0.4} & \meansd{74.2}{0.4} & \meansd{81.4}{0.1} \\%
            100 & \meansd{69.3}{0.4} & \meansd{72.4}{0.4} & \meansd{77.9}{0.4} & \meansd{84.5}{0.2} & \meansd{69.1}{0.4} & \meansd{84.2}{0.2} & \meansd{86.5}{0.2} & \meansd{87.7}{0.1} & \meansd{87.5}{0.3} & \meansd{86.4}{0.4} & \meansd{93.3}{0.1} & \meansd{89.2}{0.2} & \meansd{76.6}{0.3} & \meansd{80.4}{0.4} & \meansd{74.7}{0.4} & \meansd{81.3}{0.2} \\ %
            200 & \meansd{67.0}{0.3} & \meansd{70.9}{0.2} & \meansd{76.4}{0.4} & \meansd{83.1}{0.2} & \meansd{66.9}{0.5} & \meansd{83.0}{0.3} & \meansd{85.1}{0.1} & \meansd{86.5}{0.1} & \meansd{86.7}{0.2} & \meansd{85.2}{0.2} & \meansd{92.5}{0.1} & \meansd{87.0}{0.3} & \meansd{74.8}{0.1} & \meansd{78.0}{0.4} & \meansd{73.3}{0.1} & \meansd{79.8}{0.1} \\
            400 & \meansd{63.3}{0.3} & \meansd{67.9}{0.2} & \meansd{72.9}{0.4} & \meansd{81.8}{0.2} & \meansd{61.9}{0.2} & \meansd{81.7}{0.2} & \meansd{83.6}{0.1} & \meansd{84.9}{0.2} & \meansd{85.7}{0.1} & \meansd{82.9}{0.2} & \meansd{91.5}{0.1} & \meansd{83.7}{0.1} & \meansd{72.0}{0.2} & \meansd{74.4}{0.5} & \meansd{71.5}{0.4} & \meansd{77.3}{0.1} \\ 
            \midrule
            & \multicolumn{16}{c}{CIFAR100-C (fixed seed: 48.2)} \\
            \cmidrule(lr){2-17}
            Batch Size & \multicolumn{3}{c}{Noise} & \multicolumn{4}{c}{Blur} & \multicolumn{4}{c}{Weather} & \multicolumn{4}{c}{Digital} & \multirow{2.5}{*}{Avg.} \\
            \cmidrule(lr){2-4} \cmidrule(lr){5-8} \cmidrule(lr){9-12} \cmidrule(lr){13-16}
            & Gauss. & Shot & Impul. & Defoc. & Glass & Motion & Zoom & Snow & Frost & Fog & Brit. & Contr. & Elastic & Pixel & JPEG & \\
            \midrule
            20 & \meansd{35.9}{0.1} & \meansd{37.8}{0.6} & \meansd{42.3}{0.3} & \meansd{52.6}{0.2} & \meansd{36.8}{0.2} & \meansd{51.6}{0.4} & \meansd{55.7}{0.2} & \meansd{54.5}{0.2} & \meansd{54.6}{0.2} & \meansd{51.9}{0.4} & \meansd{64.0}{0.0} & \meansd{54.6}{0.3} & \meansd{43.3}{0.3} & \meansd{48.6}{0.3} & \meansd{41.9}{0.2} & \meansd{48.4}{0.1} \\
            50 & \meansd{38.5}{0.2} & \meansd{40.9}{0.2} & \meansd{46.3}{0.3} & \meansd{55.9}{0.3} & \meansd{38.7}{0.5} & \meansd{55.4}{0.2} & \meansd{60.0}{0.2} & \meansd{58.7}{0.2} & \meansd{58.7}{0.2} & \meansd{56.0}{0.3} & \meansd{69.0}{0.1} & \meansd{58.9}{0.5} & \meansd{45.5}{0.3} & \meansd{50.6}{0.2} & \meansd{44.5}{0.2} & \meansd{51.8}{0.1} \\ %
            100 & \meansd{37.2}{0.2} & \meansd{39.1}{0.2} & \meansd{45.3}{0.4} & \meansd{55.2}{0.2} & \meansd{35.9}{0.2} & \meansd{54.6}{0.1} & \meansd{59.5}{0.2} & \meansd{58.3}{0.2} & \meansd{58.1}{0.1} & \meansd{55.2}{0.1} & \meansd{69.0}{0.0} & \meansd{57.1}{0.3} & \meansd{43.7}{0.1} & \meansd{48.3}{0.5} & \meansd{43.6}{0.1} & \meansd{50.7}{0.1} \\
            200 & \meansd{33.8}{0.3} & \meansd{36.4}{0.1} & \meansd{43.1}{0.2} & \meansd{53.2}{0.1} & \meansd{32.2}{0.1} & \meansd{52.8}{0.2} & \meansd{58.0}{0.1} & \meansd{56.8}{0.3} & \meansd{56.6}{0.2} & \meansd{53.1}{0.1} & \meansd{67.6}{0.1} & \meansd{52.4}{0.1} & \meansd{40.9}{0.0} & \meansd{43.6}{0.2} & \meansd{41.8}{0.1} & \meansd{48.2}{0.1} \\
            400 & \meansd{29.9}{0.3} & \meansd{33.6}{0.1} & \meansd{40.4}{0.2} & \meansd{51.3}{0.2} & \meansd{28.7}{0.1} & \meansd{50.5}{0.2} & \meansd{56.2}{0.0} & \meansd{54.6}{0.3} & \meansd{55.1}{0.1} & \meansd{51.0}{0.1} & \meansd{65.8}{0.0} & \meansd{47.5}{0.1} & \meansd{37.7}{0.1} & \meansd{38.1}{0.1} & \meansd{39.9}{0.1} & \meansd{45.4}{0.2} \\

            \midrule
            & \multicolumn{16}{c}{ImageNet-C (fixed seed: 36.9)} \\
             \cmidrule(lr){2-17}
            Batch Size & \multicolumn{3}{c}{Noise} & \multicolumn{4}{c}{Blur} & \multicolumn{4}{c}{Weather} & \multicolumn{4}{c}{Digital} & \multirow{2.5}{*}{Avg.} \\
            \cmidrule(lr){2-4} \cmidrule(lr){5-8} \cmidrule(lr){9-12} \cmidrule(lr){13-16}
            & Gauss. & Shot & Impul. & Defoc. & Glass & Motion & Zoom & Snow & Frost & Fog & Brit. & Contr. & Elastic & Pixel & JPEG & \\
            \midrule
            16 & \meansd{22.0}{0.3} & \meansd{23.6}{0.3} & \meansd{21.2}{0.2} & \meansd{36.9}{0.4} & \meansd{25.6}{0.4} & \meansd{36.7}{0.3} & \meansd{39.1}{0.5} & \meansd{33.2}{0.1} & \meansd{31.5}{0.7} & \meansd{42.6}{0.1} & \meansd{56.6}{0.2} & \meansd{50.2}{0.8} & \meansd{34.3}{0.4} & \meansd{35.6}{0.2} & \meansd{30.2}{0.7} & \meansd{34.6}{0.2} \\
            32 & \meansd{24.0}{0.4} & \meansd{26.2}{0.2} & \meansd{23.3}{0.5} & \meansd{37.9}{0.3} & \meansd{27.1}{0.4} & \meansd{39.0}{0.2} & \meansd{41.2}{0.6} & \meansd{34.8}{0.3} & \meansd{35.0}{0.7} & \meansd{44.4}{0.2} & \meansd{58.1}{0.2} & \meansd{50.1}{1.1} & \meansd{37.3}{0.1} & \meansd{37.1}{0.3} & \meansd{32.2}{0.6} & \meansd{36.5}{0.1} \\
            64 & \meansd{25.3}{0.5} & \meansd{26.7}{0.3} & \meansd{24.0}{0.2} & \meansd{38.4}{0.4} & \meansd{28.1}{0.5} & \meansd{40.0}{0.5} & \meansd{42.2}{0.4} & \meansd{35.5}{0.3} & \meansd{36.3}{0.2} & \meansd{45.1}{0.4} & \meansd{58.6}{0.2} & \meansd{48.0}{1.4} & \meansd{38.9}{0.3} & \meansd{37.3}{0.2} & \meansd{32.1}{0.4} & \meansd{37.1}{0.2} \\
            128 & \meansd{25.3}{0.2} & \meansd{26.4}{0.5} & \meansd{24.0}{0.1} & \meansd{38.4}{0.1} & \meansd{28.0}{0.5} & \meansd{39.7}{0.4} & \meansd{41.7}{0.5} & \meansd{35.1}{0.3} & \meansd{36.4}{0.2} & \meansd{45.0}{0.3} & \meansd{58.6}{0.2} & \meansd{45.7}{1.9} & \meansd{39.1}{0.5} & \meansd{36.8}{0.3} & \meansd{31.8}{0.7} & \meansd{36.8}{0.1} \\
            \bottomrule
        \end{tabular}
    }
    }
\end{table}

\renewcommand{\meansd}[2]{#1 $\pm$ #2}

\subsection{Decomposition of GT-IM}
Table~\ref{tab:abl_gtim} decomposes the GT-IM objective of Eq.~\eqref{eq:tsim} on CIFAR10-C and CIFAR100-C. Variants (a) to (d) use CLIP only, and variant (e) adds the DINO\,v2 logit ensemble. OLA is disabled in all rows. The marginal term alone raises the accuracy on CIFAR10-C from 61.9 to 67.9 but lowers it on CIFAR100-C from 37.0 to 35.6, so it cannot replace per-sample sharpening. Combining the two terms gives 74.4 and 44.0, which exceeds either term alone on both datasets. The adaptive gate adds 2.5\% on CIFAR10-C and 1.2\% on CIFAR100-C, and the ensemble adds a further 2.3\% on both datasets. These results show that the conditional and marginal terms are complementary and that the gate and the ensemble each contribute on top of them. ImageNet-C is not included in this decomposition.

\begin{table}[t]
\centering
\scriptsize
\caption{Decomposition of GT-IM on CLIP+TENT. Cond.: conditional-entropy term, Marg.: marginal-entropy term, Gate: adaptive confidence gate, Ens.: DINO\,v2 logit ensemble. OLA is not used in any row.}
\vspace{5pt}
\label{tab:abl_gtim}
\setlength{\tabcolsep}{5pt}
\renewcommand{\arraystretch}{0.95}
\begin{tabular}{@{}lcccccc@{}}
\toprule
Variant & Cond. & Marg. & Gate & Ens. & CIFAR10-C & CIFAR100-C \\
\midrule
(a) conditional only (= TENT) & \checkmark &            &            &            & 61.9 & 37.0 \\
(b) marginal only             &            & \checkmark &            &            & 67.9 & 35.6 \\
(c) both, no gate             & \checkmark & \checkmark &            &            & 74.4 & 44.0 \\
(d) + gate (full GT-IM)       & \checkmark & \checkmark & \checkmark &            & 76.9 & 45.2 \\
(e) + ensemble                & \checkmark & \checkmark & \checkmark & \checkmark & \textbf{79.2} & \textbf{47.5} \\
\bottomrule
\end{tabular}
\end{table}

\subsection{Comparison with baselines using the same auxiliary backbone}
A natural question is whether the gain of \textsc{CDIM} comes only from the additional DINO\,v2 backbone. Table~\ref{tab:fair} therefore adds the DINO\,v2 logit ensemble of Eq.~\eqref{eq:ensemble} to TENT and BATCLIP on CIFAR100-C and ImageNet-C. The ensemble improves TENT by 3.0 and 3.5\% and BATCLIP by 2.4 and 2.5\% on the two datasets. \textsc{CDIM} still exceeds the strongest of these variants, BATCLIP with the DINO\,v2 ensemble, by 3.6\% on CIFAR100-C and 4.3\% on ImageNet-C. These results suggest that the DINO\,v2 backbone alone does not account for the improvement of \textsc{CDIM}.

\begin{table}[t]
\centering
\scriptsize
\caption{Accuracy (\%) of TENT and BATCLIP with and without the DINO\,v2 logit ensemble (default setting: CLIP ViT-B/16 and DINO\,v2-base).}
\vspace{5pt}
\label{tab:fair}
\setlength{\tabcolsep}{8pt}
\renewcommand{\arraystretch}{0.95}
\begin{tabular}{@{}lcc@{}}
\toprule
Method & CIFAR100-C & ImageNet-C \\
\midrule
TENT (CLIP only)                & 37.0 & 25.2 \\
TENT + DINO\,v2 ensemble          & 40.0 & 28.7 \\
BATCLIP (CLIP only)             & 42.2 & 30.1 \\
BATCLIP + DINO\,v2 ensemble       & 44.6 & 32.6 \\
\textsc{CDIM} (Ours)            & \textbf{48.2} & \textbf{36.9} \\
\bottomrule
\end{tabular}
\end{table}

\subsection{Sensitivity to Batch Size and Random Seed}
\label{sub:bat}
Table~\ref{tab:batch} reports the mean (std) over five random seeds for $B \in \{20, 50, 100, 200, 400\}$ on CIFAR10-C and CIFAR100-C and for $B \in \{16, 32, 64, 128\}$ on ImageNet-C. At the default batch size (200 for CIFAR and 64 for ImageNet-C), the five-seed mean matches the fixed-seed result on CIFAR10-C (79.8 vs.\ 79.8) and CIFAR100-C (48.2 vs.\ 48.2) and differs by $+0.2$ on ImageNet-C (37.1 vs.\ 36.9). 
 The seed protocol therefore does not change the headline numbers. Accuracy is not monotone in $B$. For $B = 20, 50, 100, 200, 400$, it is 79.5, 81.4, 81.3, 79.8, and 77.3 on CIFAR10-C and 48.4, 51.8, 50.7, 48.2, and 45.4 on CIFAR100-C. For $B = 16, 32, 64, 128$, it is 34.6, 36.5, 37.1, and 36.8 on ImageNet-C.
 Accuracy is highest at $B \in [50, 100]$ on CIFAR10-C and CIFAR100-C and at $B = 64$ on ImageNet-C. The default batch size of the CIFAR experiments is therefore not the best setting for \textsc{CDIM}: $B = 50$ is 1.6\% higher on CIFAR10-C and 3.6\% higher on CIFAR100-C. The main experiments use a fixed dataset seed and batch sizes of 200, 200, and 64 for CIFAR10-C, CIFAR100-C, and ImageNet-C to follow the protocol of~\cite{maharana2025batclip}. The baselines are not re-tuned for batch size, so Table~\ref{tab:batch} should not be read as a comparison with them.
Two effects are mixed in this trend. First, $B$ determines the quality of the batch-level estimates, such as the batch-mean prediction $\bar{p}$ in the marginal-entropy term of GT-IM. Small batches ($B \leq 20$) give noisy estimates, which is a plausible reason for the lower accuracy at small $B$. 
Second, for a fixed test set, $B$ also determines the number of parameter updates. 
The decrease at large $B$ may therefore reflect fewer updates, a more averaged gradient, or a reduced ability to follow instance-specific distribution shifts in the online stream.

\subsection{Robustness Across Corruption Severity Levels}
\label{subsec:new-abl-severity}
 
Table~\ref{tab:severity_results} reports the full severity-level breakdown (levels 1--5) of CDIM on ImageNet-C, CIFAR10-C, and CIFAR100-C, extending the severity-5 results reported in Table~1 of the main text. 
The mean accuracy decreases monotonically with severity on all three datasets (ImageNet-C: 60.1, 55.2, 51.5, 44.8, 36.9 for levels 1–5). 
 Per-corruption accuracy is not monotone in every case. For example, elastic transform on ImageNet-C is 44.6 at level 2 and 57.1 at level 3, and snow is 49.1 at level 2 and 50.3 at level 3.

 \begin{table*}[t]
\centering
\scriptsize
\caption{CDIM accuracy (\%) across corruption severity levels 1--5 on
ImageNet-C, CIFAR10-C, and CIFAR100-C.}
\label{tab:severity_results}
\setlength{\tabcolsep}{2pt}
\renewcommand{\arraystretch}{0.95}
\begin{tabular}{@{}lc*{15}{c}c@{}}
\toprule
\multirow{2}{*}{Dataset} & \multirow{2}{*}{Severity} & \multicolumn{3}{c}{Noise} & \multicolumn{4}{c}{Blur} & \multicolumn{4}{c}{Weather} & \multicolumn{4}{c}{Digital} & \multirow{2}{*}{Mean$\uparrow$} \\
\cmidrule(lr){3-5}\cmidrule(lr){6-9}\cmidrule(lr){10-13}\cmidrule(lr){14-17}
& & \multicolumn{1}{c}{Gauss.} 
& \multicolumn{1}{c}{Shot} 
& \multicolumn{1}{c}{Impul.}
& \multicolumn{1}{c}{Defoc.}
& \multicolumn{1}{c}{Glass}
& \multicolumn{1}{c}{Motion}
& \multicolumn{1}{c}{Zoom}
& \multicolumn{1}{c}{Snow}
& \multicolumn{1}{c}{Frost}
& \multicolumn{1}{c}{Fog}
& \multicolumn{1}{c}{Brit.}
& \multicolumn{1}{c}{Contr.}
& \multicolumn{1}{c}{Elast.}
& \multicolumn{1}{c}{Pixel.}
& \multicolumn{1}{c}{JPEG}
& \\
\midrule
\multirow{5}{*}{ImageNet-C} 
& Level 1 & 60.5 & 60.0 & 56.9 & 58.3 & 58.3 & 62.2 & 54.4 & 57.7 & 58.7 & 60.8 & 66.0 & 63.4 & 60.6 & 62.1 & 61.3 & 60.1 \\
& Level 2 & 56.2 & 55.1 & 51.8 & 54.1 & 53.9 & 58.0 & 48.5 & 49.1 & 51.0 & 58.9 & 65.4 & 62.5 & 44.6 & 60.5 & 59.1 & 55.2 \\
& Level 3 & 48.2 & 47.7 & 46.9 & 46.7 & 41.7 & 51.2 & 44.4 & 50.3 & 44.8 & 56.3 & 63.8 & 58.9 & 57.1 & 56.9 & 57.1 & 51.5 \\
& Level 4 & 38.3 & 35.7 & 36.9 & 39.0 & 35.2 & 41.9 & 40.2 & 44.9 & 42.8 & 52.9 & 61.8 & 49.9 & 50.8 & 49.7 & 51.5 & 44.8 \\
  & Level 5 & \bf{26.0} & \bf{27.9} & \bf{26.7} & \bf{31.1} & \bf{30.1} & \bf{35.9} & \bf{34.4} & \bf{42.2} & \bf{38.3} & \bf{45.6} & \bf{59.6} & \bf{34.1} & \bf{33.1} & \bf{45.1} & \bf{43.8} & \bf{36.9} \\
\midrule
\multirow{5}{*}{CIFAR10-C} 
& Level 1 & 88.0 & 90.3 & 92.9 & 94.4 & 78.0 & 92.0 & 90.8 & 92.7 & 93.5 & 94.6 & 94.8 & 94.7 & 90.0 & 93.5 & 87.1 & 91.1 \\
& Level 2 & 82.1 & 87.7 & 91.0 & 93.4 & 77.5 & 89.7 & 90.5 & 89.5 & 91.5 & 93.2 & 94.6 & 93.9 & 89.7 & 90.9 & 82.3 & 89.2 \\
& Level 3 & 74.1 & 80.3 & 88.2 & 91.7 & 76.7 & 85.6 & 89.1 & 89.4 & 89.1 & 92.5 & 94.3 & 93.4 & 87.9 & 90.3 & 79.8 & 86.8 \\
& Level 4 & 70.5 & 75.8 & 82.5 & 88.8 & 68.0 & 86.1 & 87.8 & 86.8 & 88.9 & 90.3 & 93.8 & 92.2 & 81.6 & 85.3 & 77.3 & 83.7 \\
  & Level 5 & \bf{67.8} & \bf{70.9} & \bf{76.3} & \bf{83.2} & \bf{67.1} & \bf{83.0} & \bf{85.2} & \bf{86.2} & \bf{87.0} & \bf{84.8} & \bf{92.5} & \bf{87.0} & \bf{74.9} & \bf{77.5} & \bf{73.4} & \bf{79.8} \\
\midrule
\multirow{5}{*}{CIFAR100-C} 
& Level 1 & 59.2 & 64.2 & 69.2 & 73.6 & 44.5 & 68.7 & 67.1 & 69.2 & 69.7 & 73.8 & 74.1 & 74.1 & 64.6 & 70.0 & 59.3 & 66.8 \\
& Level 2 & 49.6 & 58.8 & 64.4 & 71.7 & 44.1 & 63.5 & 66.6 & 63.1 & 65.9 & 71.2 & 73.7 & 71.5 & 64.2 & 65.9 & 51.3 & 63.0 \\
& Level 3 & 41.3 & 47.2 & 60.4 & 68.3 & 43.6 & 58.3 & 64.5 & 62.3 & 60.8 & 68.6 & 72.8 & 69.3 & 61.0 & 63.6 & 48.5 & 59.4 \\
& Level 4 & 37.2 & 42.6 & 51.0 & 62.7 & 32.6 & 58.2 & 62.3 & 59.9 & 60.6 & 64.3 & 71.3 & 66.0 & 51.5 & 57.0 & 45.9 & 54.9 \\
& Level 5 & \bf{33.9} & \bf{36.5} & \bf{42.8} & \bf{53.4} & \bf{32.5} & \bf{52.7} & \bf{58.0} & \bf{56.6} & \bf{56.5} & \bf{53.2} & \bf{67.4} & \underline{52.2} & \bf{41.1} & \bf{44.0} & \bf{41.7} & \bf{48.2} \\
\bottomrule
\end{tabular}
\vspace{-10pt}
\end{table*}
 
\subsection{Sensitivity to the EMA Momentum $\alpha$}
\label{subsec:abl-ema-momentum}

Table~\ref{tab:sens_alpha_rho}(a) presents a post-hoc sensitivity analysis of the EMA momentum $\alpha$ in the online logit adjustment (Eq.~\eqref{eq:ema}) over $\alpha \in \{0.1, 0.3, 0.5, 0.7, 0.9, 0.99\}$, with all other hyperparameters fixed.
The accuracy on CIFAR10-C varies between 79.4 and 80.3 (a spread of 0.9\%) and equals 80.3 for $\alpha\in[0.3,0.5]$.
 On CIFAR100-C, the accuracy varies between 47.1 and 48.2, with 48.0--48.2 for $\alpha\in[0.3,0.9]$ and 47.1 at $\alpha=0.99$. On ImageNet-C, the accuracy increases from 34.5 at $\alpha=0.1$ to 36.9 at $\alpha=0.9$ and is 36.1 at $\alpha=0.99$.
This pattern is consistent with the intuition that a larger label space requires a longer effective memory for the EMA estimate $\bar\pi^{(t)}_c$ (Eq.~\eqref{eq:ema}) to converge to a low-variance estimate of the predicted-class distribution. 
An excessively long memory ($\alpha = 0.99$) reacts too slowly to genuine distribution drift, which is the usual bias-variance trade-off of an exponential moving average. We report this sweep only as a sensitivity analysis and did not use it to select $\alpha$.

\subsection{Sensitivity to the Discard Ratio $\rho$}
\label{subsec:abl-discard-ratio}

Table~\ref{tab:sens_alpha_rho}(b) sweeps the discard ratio $\rho$ of the gating mask (Eq.~\eqref{eq:gate}) over $\rho\in\{0.01, 0.03, 0.05, 0.10, 0.20, 0.30\}$, with all other hyperparameters fixed. The default $\rho=0.05$ attains the highest accuracy on all three benchmarks (79.8\% on CIFAR10-C, 48.2\% on CIFAR100-C, and 36.9\% on ImageNet-C).
The accuracy at the neighboring values $\rho = 0.03$ and $\rho = 0.10$ is lower by at most 0.5\% on every benchmark. The extreme value $\rho = 0.30$ reduces the accuracy by 1.4\%, 1.5\%, and 1.2\% on CIFAR10-C, CIFAR100-C, and ImageNet-C, respectively, and $\rho = 0.01$ reduces it by 1.2\% and 1.1\% on CIFAR10-C and CIFAR100-C. On ImageNet-C, $\rho = 0.01$ and $\rho = 0.03$ give the same accuracy (0.4\% below the default) because both discard a single sample per batch at $B = 64$. The total spread over the sweep is therefore 1.4\%, 1.5\%, and 1.2\% on the three benchmarks. A small $\rho$ keeps more samples with noise-dominated entropy gradients in the sharpening term, whereas a large $\rho$ removes samples that still carry a class signal. This explanation is consistent with the observed curve on CIFAR10-C and CIFAR100-C. Note that we report the sweep over $\rho$ as a sensitivity analysis and did not use it to select $\rho$.

\begin{table}[t]
\centering
\caption{Sensitivity of \textsc{CDIM} to (a) the EMA momentum $\alpha$ (Eq.~\eqref{eq:ema}) and (b) the discard ratio $\rho$ (Eq.~\eqref{eq:gate}), measured by accuracy (\%) on CIFAR10-C, CIFAR100-C, and ImageNet-C. The default values are $\alpha=0.9$ and $\rho=0.05$ (rows 0.9 and 0.05). Bold marks the best value in each column.}
    \vspace{6pt}
\label{tab:sens_alpha_rho}
\footnotesize
\begin{subtable}[t]{0.49\linewidth}
\centering
\setlength{\tabcolsep}{3pt}
\begin{tabular}{@{}cccc@{}}
\toprule
$\alpha$ & CIFAR10-C & CIFAR100-C & ImageNet-C \\
\midrule
0.1  & 79.6 & 47.8 & 34.5 \\
0.3  & \textbf{80.3} & 48.0 & 35.0 \\
0.5  & \textbf{80.3} & 48.1 & 35.4 \\
0.7  & 80.1 & 48.1 & 36.1 \\
0.9 & 79.8 & \textbf{48.2} & \textbf{36.9} \\
0.99 & 79.4 & 47.1 & 36.1 \\
\bottomrule
\end{tabular}
\caption{EMA momentum $\alpha$}
\label{tab:sens_alpha}
\end{subtable}%
\hfill%
\begin{subtable}[t]{0.49\linewidth}
\centering
\setlength{\tabcolsep}{3pt}
\begin{tabular}{@{}cccc@{}}
\toprule
$\rho$ & CIFAR10-C & CIFAR100-C & ImageNet-C \\
\midrule
0.01 & 78.6 & 47.1 & 36.5 \\
0.03 & 79.4 & 47.9 & 36.5 \\
0.05 & \textbf{79.8} & \textbf{48.2} & \textbf{36.9} \\
0.10 & 79.3 & 47.8 & 36.6 \\
0.20 & 78.8 & 47.2 & 36.1 \\
0.30 & 78.4 & 46.7 & 35.7 \\
\bottomrule
\end{tabular}
\caption{Discard ratio $\rho$}
\label{tab:sens_rho}
\end{subtable}
\end{table}

\begin{figure*}[t]
    \centering
    \includegraphics[width=0.8\textwidth]{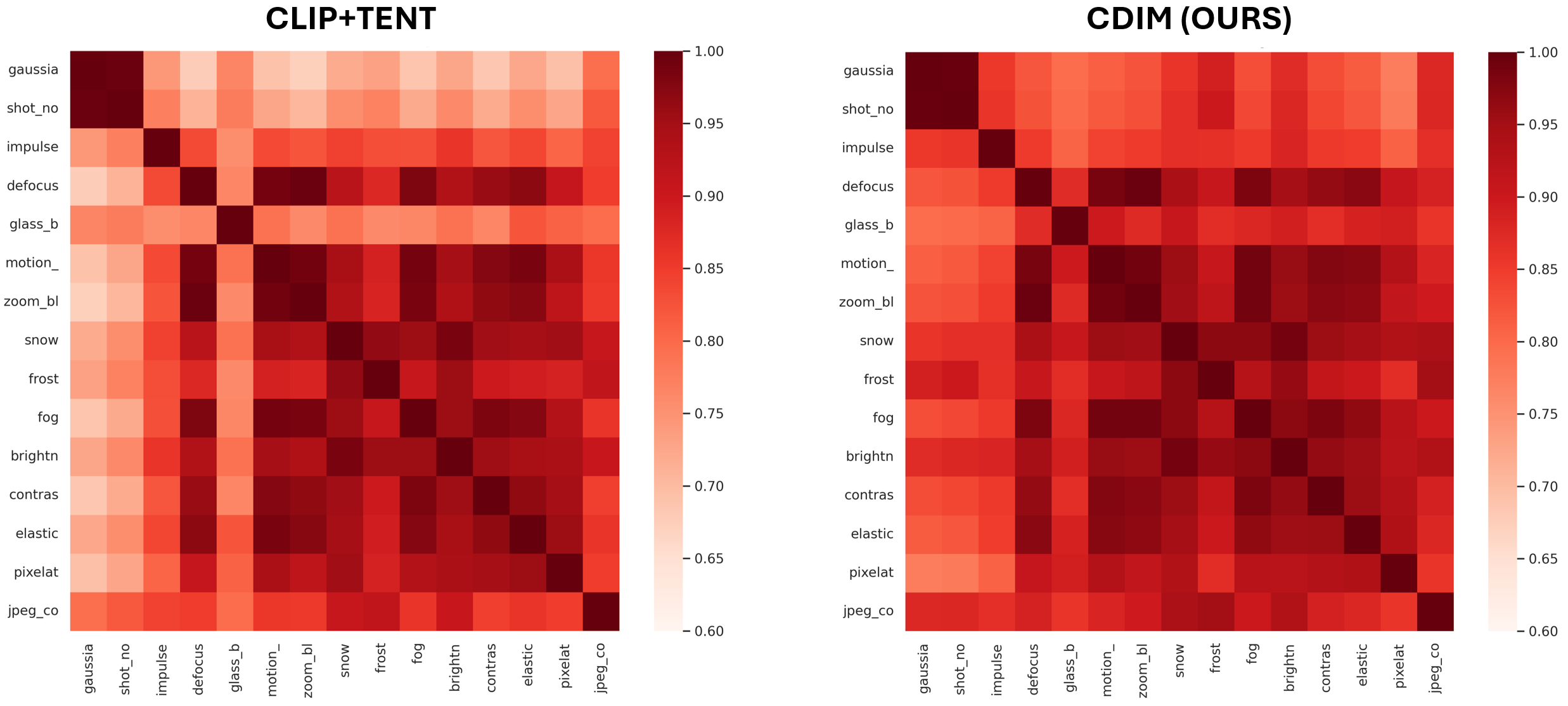}
    \caption{Visualization of the mean class-wise representation similarity across different corruptions on CIFAR10-C. \textbf{Darker colors} indicate high similarity, meaning that the model extracts consistent features regardless of the corruption type. \textbf{Lighter colors} indicate low similarity, which reveals feature distortion and representation drift caused by the distribution shift.
}
    \label{fig:heatmap}
\end{figure*}

\section{Additional Analysis}
\label{sec:new-ablation}

\begin{figure*}[t]
\centering
\begin{subtable}[t]{0.48\linewidth}
  \centering
  \includegraphics[width=1.1\linewidth]{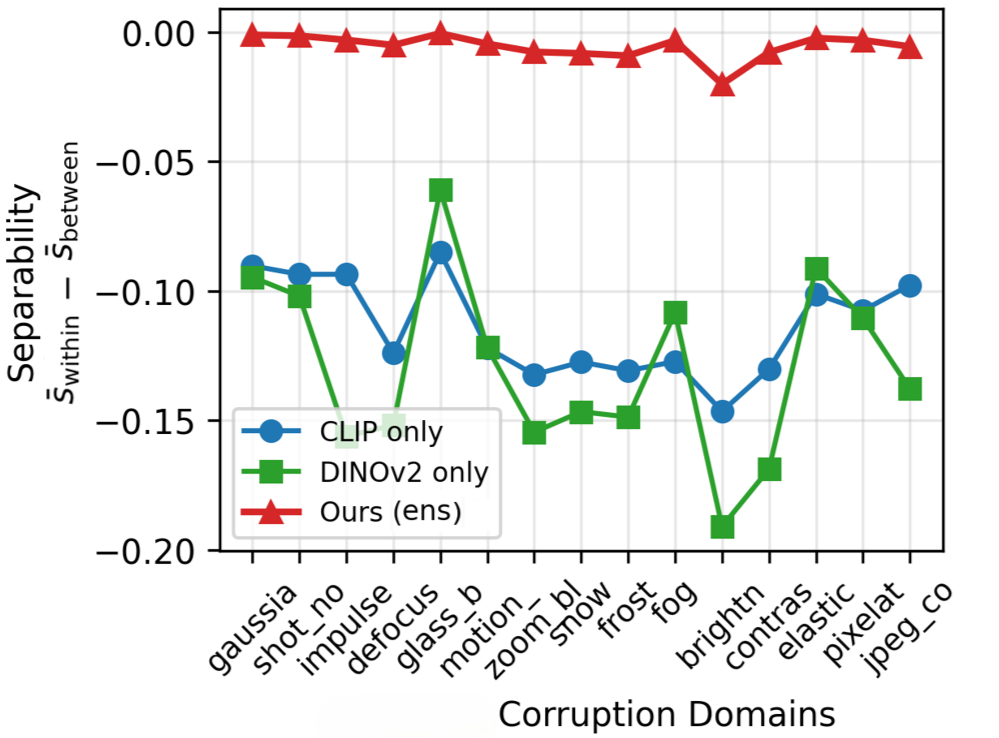}
  \caption{Class separability.}
  \label{fig:cls_sep}
\end{subtable}\hfill
\begin{subtable}[t]{0.48\linewidth}
  \centering
  \includegraphics[width=\linewidth]{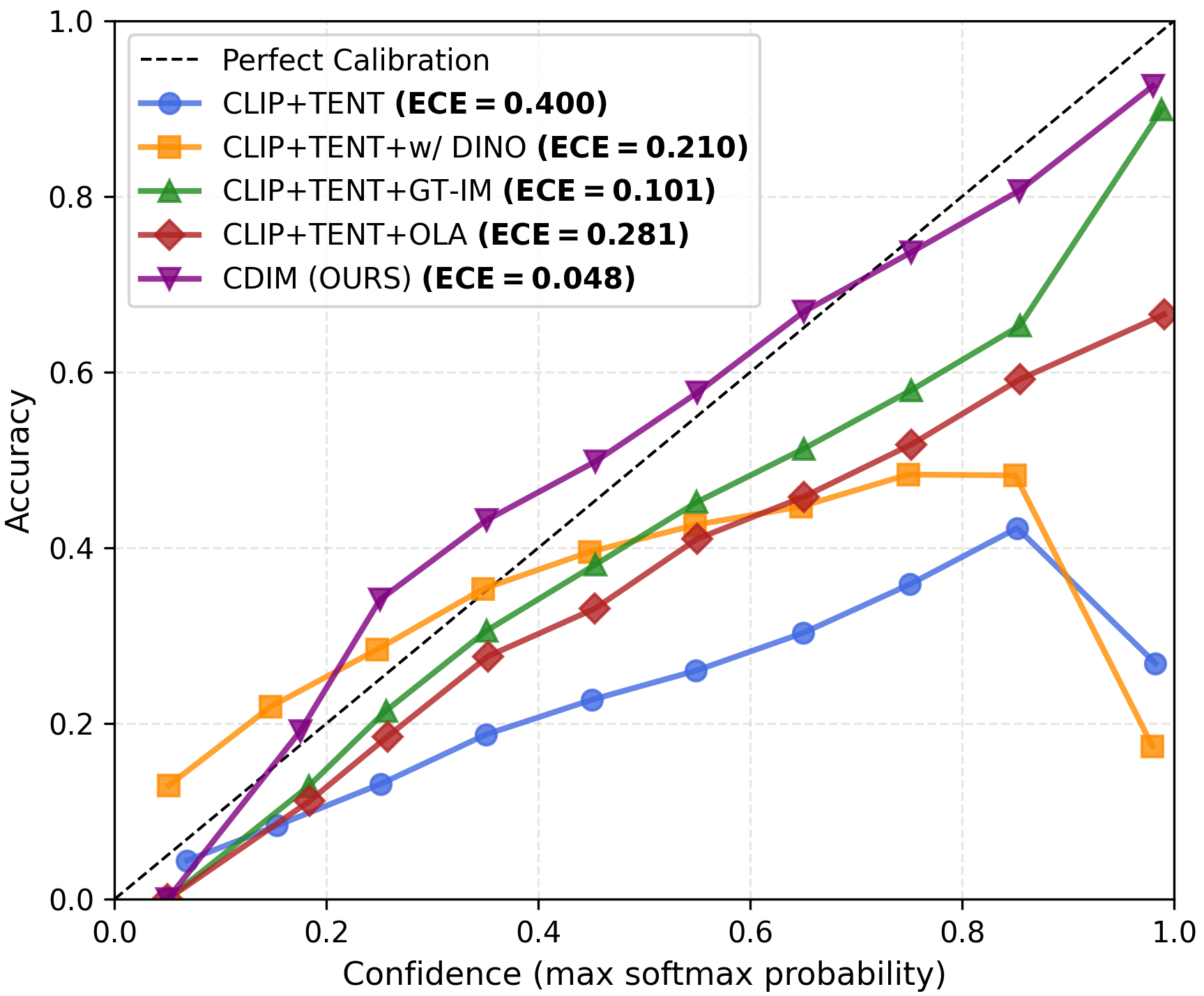}
  \caption{Additional Reliability diagram.}
  \label{fig:app_abl2_real}
\end{subtable}
\caption{(a): Class separability $\bar{s}_{\text{within}} - \bar{s}_{\text{between}}$ of the per-batch feature matrix, measured on each of the fifteen CIFAR10-C corruption types; higher values indicate better-separated classes. (b): Reliability diagram for CLIP+TENT+OLA (\textcolor{red}{red}) and CDIM (\textcolor{violet}{purple}). OLA lowers the ECE from 0.400 (CLIP+TENT) to 0.281, but residual overconfidence remains. CDIM, which adds the DINO\,v2 ensemble and GT-IM, removes most of it and reaches an ECE of 0.048.}

\end{figure*}

\paragraph{Per-corruption heatmap.}
Figure~\ref{fig:heatmap} visualizes the cross-corruption representation similarity on CIFAR10-C, computed as the average intra-class similarity between different corruption types. In the CLIP$+$TENT heatmap, adaptation to severe noise corruptions (Gaussian, shot, and impulse noise) yields representations that deviate strongly from those of the other corruptions, as indicated by lighter regions. In the \textsc{CDIM} heatmap, the intra-class similarity across corruption types is higher than under CLIP+TENT, including between the noise corruptions and the others. This indicates that \textsc{CDIM} reduces representation drift across corruption types.

\begin{figure}[t]
    \centering
    \includegraphics[width=1.0\textwidth]{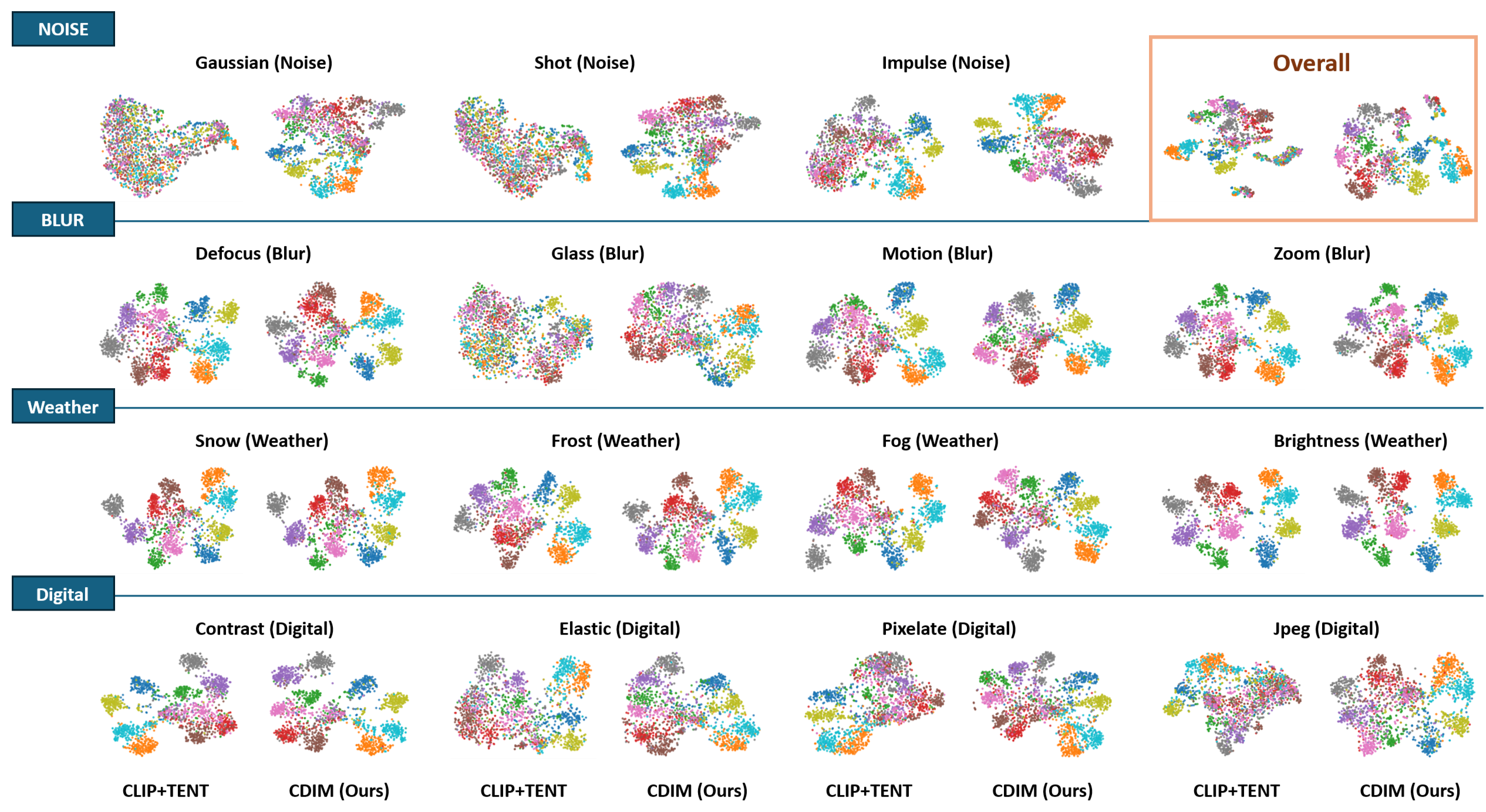}
    \caption{t-SNE visualization of CLIP+TENT and CDIM features on CIFAR10-C for all 15 corruption types, together with an overall results (top right).}
    \label{fig:app_tsne}
\end{figure}

\paragraph{Stability of the ensembled representation.}
Figure~\ref{fig:cls_sep} reports the class separability $\bar{s}_{\text{within}} - \bar{s}_{\text{between}}$ across the fifteen corruption types of CIFAR10-C, where higher values indicate better-separated classes. Here, $\bar{s}_{\text{between}}$ and $\bar{s}_{\text{within}}$ are the mean pairwise similarities of the per-batch features for sample pairs from different classes and from the same class, respectively. The ensembled representation of \textsc{CDIM} (red) varies by less than $0.02$ across corruption types, whereas the CLIP-only (blue) and DINO v2-only (green) representations vary by about $0.06$ and $0.13$, respectively.
This stability is consistent with the cross-covariance construction of Section~\ref{sec:fusion}, in which averaging the predictions of the two backbones may reduce sensitivity to corruption-vulnerable directions.

\paragraph{Comprehensive t-SNE Visualizations.}
Due to space constraints, Figure~\ref{fig:tsne} in Section~\ref{sec:experiment} displays only one representative t-SNE visualization per corruption category. Figure~\ref{fig:app_tsne} provides the t-SNE visualizations for all individual corruption types, comparing CDIM with the CLIP+TENT baseline.

\section{Additional Experiments}

\subsection{Evaluation on Domain Generalization Benchmarks}
\label{add_domain}
In this section, we evaluate \textsc{CDIM} on commonly used domain generalization (DG) datasets: VisDA-C, OfficeHome, PACS, and VLCS. We follow the DG protocol of BATCLIP~\cite{maharana2025batclip} and take the baseline results from its table for comparison. \textsc{CDIM} is compared with zero-shot CLIP (ZS), TENT~\cite{wang2020tent}, WATT~\cite{osowiechi2024watt}, and BATCLIP~\cite{maharana2025batclip}. The full results are given in Table~\ref{tab:dg-benchmarks}.
On VisDA-C (simulated and video shifts), \textsc{CDIM} reaches the highest accuracy of 86.0\% (3D) and 85.5\% (YT), exceeding recent baselines such as WATT and BATCLIP. On OfficeHome, PACS, and VLCS, where vanilla CLIP often struggles under texture and style shifts, \textsc{CDIM} improves over zero-shot CLIP in all 14 domains, with notable gains on Clipart in OfficeHome (68.0\%, $+4.0$ over ZS), SUN09 in VLCS (77.1\%, $+3.7$), and Sketch in PACS (89.2\%, $+3.0$). \textsc{CDIM} is the best or tied for the best in 10 of the 14 domains. It is not the best on Product in OfficeHome (87.7\% vs.\ 88.5\% for BATCLIP), Art in PACS (97.7\% vs.\ 98.1\% for TENT), LabelMe in VLCS (68.6\% vs.\ 68.9\% for BATCLIP), and VOC2007 in VLCS (85.4\% vs.\ 85.6\% for BATCLIP).
\begin{table*}[t]
\centering
\scriptsize
\caption{Domain generalization results across VisDA-C, OfficeHome, PACS, and VLCS. We denote Top-1/2 by \textbf{bold} and \underline{underline}, respectively.}
\label{tab:dg-benchmarks}
\setlength{\tabcolsep}{5pt}
\renewcommand{\arraystretch}{0.95}
\begin{tabular}{@{}llccccc@{}}
\toprule
Dataset & Domain & ZS & TENT  & WATT & BATCLIP & CDIM (Ours) \\
\midrule
\multirow{2}{*}{VisDA-C}
& 3D (trainset) &84.4 &84.9 &85.4 & 85.3 & \textbf{86.0} \\
& YT (valset) &84.5 &84.7 &84.7  & 84.9 & \textbf{85.5} \\
\midrule
\multirow{4}{*}{OfficeHome}
& Art & 78.3 & 79.4 & 80.4 & 79.9 & \textbf{80.7} \\
& Clipart & 64.0 & 63.2 & 66.9 & 66.4 & \textbf{68.0} \\
& Product & 87.1 & 88.4 & 87.5 & \textbf{88.5} & \underline{87.7} \\
& Real World & 89.0 & 89.6 & 90.0 & 89.7 & \textbf{90.1} \\
\midrule
\multirow{4}{*}{PACS}
& Art & 97.2 & \textbf{98.1} & \underline{97.8} & 97.6 & 97.7 \\
& Cartoon & 97.4 & \textbf{97.7} & 97.5 & 97.5 & \textbf{97.7} \\
& Photo & 99.6 & 99.6 & 99.5 & \textbf{99.7} & \underline{99.7} \\
& Sketch & 86.2 & 88.8 & 88.7 & 87.8 & \textbf{89.2} \\
\midrule
\multirow{4}{*}{VLCS}
& Caltech101 & 99.4 & 99.4  & 99.4 & \textbf{99.5} & \textbf{99.5} \\
& LabelMe    & 68.2 & 68.1 & 68.5 & \textbf{68.9} & \underline{68.6} \\
& SUN09      & 73.4 & 73.4  & 74.7 & 75.2 & \textbf{77.1} \\
& VOC2007    & 84.8 & 84.8  & 84.0 & \textbf{85.6} & \underline{85.4} \\
\bottomrule
\end{tabular}
\vspace{-8pt}
\end{table*}

\begin{table}[t]%{0.48\linewidth}
\centering
\caption{Comparison on ImageNet distribution-shift variants (ImageNet-A, -V2, -R, and -Sketch) with CLIP ViT-B/16.}
    \vspace{5pt}
\scriptsize
\setlength{\tabcolsep}{3pt}
\renewcommand{\arraystretch}{1}
\begin{tabular}{@{}lcccc@{}}
\toprule
Method & ImageNet-A & ImageNet-V2 & ImageNet-R & ImageNet-Sketch \\
\midrule
CLIP~\cite{radford2021learning}     & 49.2 & 60.4 & 72.7 & 44.9 \\
CLIPArTT~\cite{a2025clipartt} & 49.6 & 60.5 & 72.8 & 45.0 \\
TDA~\cite{karmanov2024efficient}      & 51.0 & 61.2 & 73.9 & 46.4 \\
TENT~\cite{wang2020tent}     & 51.9 & 61.0 & 77.0 & 45.4 \\
ETA~\cite{niu2022efficient}      & 52.0 & 61.0 & 77.4 & 46.8 \\
Mint~\cite{bao2026mint}     & 54.7 & 62.6 & 78.1 & 48.4 \\
BATCLIP~\cite{maharana2025batclip}     &48.7	&60.5	&72.6	&39.9 \\
\textbf{CDIM (Ours)}     & \bf{55.5} & \bf{63.3} & \bf{79.2} & \bf{50.1} \\
\bottomrule
\end{tabular}
\label{tab:mulimg}
\end{table}
\subsection{Robustness to Natural Distribution Shifts and Natural Adversarial Examples}
\label{sec:appendix_imagenet_variants}
To assess robustness beyond synthetic corruption benchmarks, we evaluate \textsc{CDIM} on four ImageNet variants that capture natural distribution shifts and natural adversarial examples: ImageNet-A, -V2, -R, and -Sketch (Table~\ref{tab:variation}(d)). For a broader comparison with existing methods, Table~\ref{tab:mulimg} follows the setting of Mint~\cite{bao2026mint} and reuses its reported baseline results alongside ours. \textsc{CDIM} achieves the best accuracy among the compared CLIP-based online TTA methods on all four datasets. It outperforms the strongest baseline, Mint~\cite{bao2026mint}, by 0.8\%, 0.7\%, 1.1\%, and 1.7\% on ImageNet-A, -V2, -R, and -Sketch (-S), respectively. The margins against CLIPArTT~\cite{a2025clipartt} and TDA~\cite{karmanov2024efficient} are larger, at least 2.8\% and 2.1\%, respectively. Against BATCLIP~\cite{maharana2025batclip}, the margins are 6.8\%, 2.8\%, 6.6\%, and 10.2\% on the four datasets. These results indicate that the robustness of CDIM extends beyond synthetic corruption benchmarks to natural distribution shifts.

\subsection{Additional Comparison Results on Continual TTA}
Due to space constraints, Table~\ref{tab:continual} in Section~\ref{sec:experiment} reports only the mean accuracy for the continual TTA setting. Table~\ref{tab:app_continual} provides the accuracy for each corruption type together with the mean.

\begin{table*}[t]
\centering
\scriptsize
\caption{Continual TTA results with ViT-B/16, including per-corruption accuracy. Top-1 and Top-2 are shown in \textbf{bold} and \underline{underline}, respectively.}
\label{tab:app_continual}
\setlength{\tabcolsep}{2pt}
\renewcommand{\arraystretch}{0.95}
\begin{tabular}{@{}ll*{15}{c}c@{}}
\toprule
\multicolumn{2}{l}{\multirow{2}{*}{Method}} & \multicolumn{3}{c}{Noise} & \multicolumn{4}{c}{Blur} & \multicolumn{4}{c}{Weather} & \multicolumn{4}{c}{Digital} & \multicolumn{1}{c}{\multirow{2}{*}{Mean$\uparrow$}} \\
\cmidrule(lr){3-5}\cmidrule(lr){6-9}\cmidrule(lr){10-13}\cmidrule(lr){14-17}
\multicolumn{2}{l}{} 
& \multicolumn{1}{c}{Gauss.} 
& \multicolumn{1}{c}{Shot} 
& \multicolumn{1}{c}{Impul.}
& \multicolumn{1}{c}{Defoc.}
& \multicolumn{1}{c}{Glass}
& \multicolumn{1}{c}{Motion}
& \multicolumn{1}{c}{Zoom}
& \multicolumn{1}{c}{Snow}
& \multicolumn{1}{c}{Frost}
& \multicolumn{1}{c}{Fog}
& \multicolumn{1}{c}{Brit.}
& \multicolumn{1}{c}{Contr.}
& \multicolumn{1}{c}{Elast.}
& \multicolumn{1}{c}{Pixel.}
& \multicolumn{1}{c}{JPEG}
& \multicolumn{1}{c}{} \\
\midrule
\multirow{8}{*}{\rotatebox{90}{ImageNet-C}} & TENT~\cite{wang2020tent} & 5.1 & 0.3 & 0.2 & 19.5 & 16.1 & 25.9 & 25.5 & 32.0 & 30.5 & 37.6 & 52.8 & 25.6 & 14.6 & 35.0 & 32.2 & 23.5  \\
 & CoTTA~\cite{wang2022continual} & 17.6 & 18.3 & 18.8 & 17.9 & 19.2 & 16.3 & 28.7 & 19.0 & 16.9 & 17.5 & 41.5 & 24.2 & 26.4 & 20.8 & 25.4 & 21.9  \\
 & EATA~\cite{niu2022efficient} & \underline{22.8} & \underline{26.9} & \underline{27.0} & \underline{25.7} & \underline{27.0} & 30.7 & \underline{31.6} & \underline{37.1} & 37.2 & \textbf{43.3} & \textbf{54.5} & \underline{33.4} & \underline{34.6} & \underline{39.9} & \underline{39.6} & \underline{34.1} \\
 & SAR~\cite{niu2023towards} & 19.5 & 21.6 & 18.2 & 25.3 & 21.3 & \textbf{32.9} & 22.8 & 32.5 & 28.8 & 42.4 & 48.6 & 28.6 & 12.5 & 39.8 & 37.8 & 28.8 \\
 & COME~\cite{zhang2024come} & 22.5 & 26.8 & 26.7 & 24.5 & 26.2 & 29.9 & 31.3 & 36.5 & \underline{37.3} & \underline{43.0} & \underline{54.0} & \textbf{33.7} & \textbf{35.8} & \textbf{40.4} & \textbf{40.1} & 33.9 \\
 & BATCLIP~\cite{maharana2025batclip} & 19.4 & 20.7 & 19.2 & 16.7 & 16.1 & 17.4 & 16.2 & 19.7 & 20.1 & 18.8 & 31.1 & 12.2 & 14.2 & 17.4 & 17.1 & 18.4 \\
 & Mint~\cite{bao2026mint} & 19.4 & 20.6 & 21.8 & 16.4 & 15.1 & 23.6 & 19.9 & 27.7 & 26.4 & 31.0 & 49.3 & 20.0 & 16.8 & 36.4 & 35.6 & 25.3 \\
 & \textbf{CDIM (Ours)} & \textbf{25.9} & \textbf{31.3} & \textbf{31.6} & \textbf{28.8} & \textbf{29.5} & \underline{31.6} & \textbf{32.4} & \textbf{37.5} & \textbf{38.7} & 39.7 & \underline{54.0} & 29.8 & 34.0 & 37.8 & 36.0 & \textbf{34.6} \\
\midrule
\multirow{8}{*}{\rotatebox{90}{CIFAR10-C}} & TENT~\cite{wang2020tent} & 37.3 & 36.8 & 49.6 & 69.5 & 43.9 & 70.4 & 73.9 & 73.9 & 76.1 & 70.6 & 83.7 & 71.5 & 57.7 & 46.2 & 59.4 & 61.4 \\
& CoTTA~\cite{wang2022continual} & 27.4 & 32.0 & 35.7 & 66.4 & 49.7 & 67.5 & 73.2 & 76.4 & 76.7 & 64.8 & 87.6 & 59.6 & 58.4 & 44.9 & 60.2 & 58.7 \\
 & EATA~\cite{niu2022efficient} & 42.5 & 47.0 & 57.8 & 73.0 & 42.7 & 70.9 & 75.3 & 75.8 & 79.1 & 72.3 & 85.6 & 64.8 & 56.3 & 48.7 & \underline{60.8} & 63.5 \\
 & SAR~\cite{niu2023towards} & 32.4 & 30.6 & 48.5 & \underline{77.0} & 41.0 & 56.9 & \underline{76.3} & \underline{82.3} & \underline{83.5} & \underline{81.9} & \underline{88.8} & \underline{80.0} & 62.7 & 48.0 & 50.8 & 62.7 \\
 & COME~\cite{zhang2024come} & 42.3 & 46.9 & 57.5 & 72.9 & 42.5 & 71.0 & 75.4 & 75.8 & 79.0 & 72.4 & 85.5 & 64.8 & 56.2 & \underline{48.8} & 60.5 & 63.4 \\
 & BATCLIP~\cite{maharana2025batclip} & \underline{62.6} & 62.2 & 58.0 & 63.0 & \underline{52.2} & 59.8 & 62.6 & 60.2 & 61.4 & 57.1 & 60.7 & 57.6 & 47.0 & 48.6 & 43.4 & 57.1 \\
 & Mint~\cite{bao2026mint} & 62.2 & \underline{65.8} & \underline{62.8} & 69.0 & 44.3 & \underline{73.9} & 74.4 & 77.9 & 78.1 & 75.8 & 88.2 & 71.8 & \underline{63.6} & 43.9 & 59.7 & \underline{67.4} \\
 & \textbf{CDIM (Ours)} & \textbf{67.8} & \textbf{75.3} & \textbf{77.1} & \textbf{81.7} & \textbf{68.6} & \textbf{81.1} & \textbf{86.0} & \textbf{85.9} & \textbf{87.6} & \textbf{85.9} & \textbf{92.0} & \textbf{87.7} & \textbf{76.4} & \textbf{78.2} & \textbf{75.4} & \textbf{80.4} \\
\midrule
\multirow{8}{*}{\rotatebox{90}{CIFAR100-C}} & TENT~\cite{wang2020tent} & 13.8 & 15.8 & 15.6 & 36.3 & 16.5 & 39.5 & 46.7 & 44.6 & 42.5 & 27.3 & 60.4 & 31.4 & 34.0 & 18.6 & 32.6 & 31.7  \\
 & CoTTA~\cite{wang2022continual} & 15.6 & 17.4 & 19.5 & 36.0 & 19.2 & 36.4 & 44.1 & 41.4 & 39.1 & 20.5 & 54.9 & 24.3 & 29.4 & 18.0 & 28.2 & 29.6  \\
 & EATA~\cite{niu2022efficient} & 26.7 & 35.4 & 40.6 & \underline{49.9} & \underline{28.3} & \underline{50.0} & \underline{57.1} & \textbf{54.7} & \textbf{56.1} & \textbf{51.2} & \textbf{66.4} & \textbf{52.7} & \underline{37.0} & \textbf{40.9} & \textbf{39.7} & \underline{45.8} \\
 & SAR~\cite{niu2023towards} & 27.2 & \underline{36.4} & \underline{41.4} & 48.1 & \underline{28.3} & 35.8 & 25.8 & \underline{54.6} & 53.5 & 42.4 & 38.8 & 30.1 & 33.6 & 37.8 & 32.1 & 37.7 \\
 & COME~\cite{zhang2024come} & \underline{27.6} & 32.8 & 36.5 & 43.8 & 22.6 & 44.2 & 53.3 & 51.7 & 53.7 & \underline{48.6} & \underline{63.9} & \underline{48.0} & 35.5 & 36.9 & \underline{37.2} & 42.4 \\
 & BATCLIP~\cite{maharana2025batclip} & 24.9 & 25.9 & 23.3 & 32.0 & 20.2 & 30.1 & 33.6 & 29.8 & 30.6 & 27.1 & 37.6 & 29.1 & 21.7 & 18.8 & 19.9 & 27.0 \\
 & Mint~\cite{bao2026mint} & 21.4 & 33.1 & 25.8 & 35.0 & 15.3 & 37.3 & 43.4 & 46.0 & 45.2 & 41.4 & 60.2 & 36.8 & 30.0 & 16.9 & 29.7 & 34.5  \\
 & \textbf{CDIM (Ours)} & \textbf{35.1} & \textbf{44.3} & \textbf{46.7} & \textbf{53.8} & \textbf{38.7} & \textbf{51.2} & \textbf{57.7} & 52.5 & \underline{54.7} & 47.0 & 57.1 & 40.6 & \textbf{40.9} & \underline{39.4} & 34.1 & \textbf{46.3} \\
\bottomrule
\end{tabular}
\vspace{-10pt}
\end{table*}

\newpage
\subsection{Additional Comparison Results on Online TTA}
Table~\ref{tab:app_online} provides the full evaluation results for the online TTA experiments discussed in Section~\ref{sec:experiment}. Table~\ref{tab:benchmark_all} in the main text lists only a subset of the baselines. Table~\ref{tab:app_online} adds DMN~\cite{zhang2024dual}, VTE~\cite{dobler2024lost}, ZERO~\cite{farina2024frustratingly}, ECALP~\cite{li2025efficient}, RoTTA~\cite{yuan2023robust}, and MEMO~\cite{zhang2022memo} to the baselines of Table 1.

\begin{table*}[t]
\centering
\scriptsize
\caption{Online TTA results with ViT-B/16, including additional baselines. Top-1 and Top-2 are shown in \textbf{bold} and \underline{underline}, respectively.}
\label{tab:app_online}
\setlength{\tabcolsep}{2pt}
\renewcommand{\arraystretch}{0.95}
\begin{tabular}{@{}ll*{15}{c}c@{}}
\toprule
\multicolumn{2}{l}{\multirow{2}{*}{Method}} & \multicolumn{3}{c}{Noise} & \multicolumn{4}{c}{Blur} & \multicolumn{4}{c}{Weather} & \multicolumn{4}{c}{Digital} & \multicolumn{1}{c}{\multirow{2}{*}{Mean$\uparrow$}} \\
\cmidrule(lr){3-5}\cmidrule(lr){6-9}\cmidrule(lr){10-13}\cmidrule(lr){14-17}
\multicolumn{2}{l}{} 
& \multicolumn{1}{c}{Gauss.} 
& \multicolumn{1}{c}{Shot} 
& \multicolumn{1}{c}{Impul.}
& \multicolumn{1}{c}{Defoc.}
& \multicolumn{1}{c}{Glass}
& \multicolumn{1}{c}{Motion}
& \multicolumn{1}{c}{Zoom}
& \multicolumn{1}{c}{Snow}
& \multicolumn{1}{c}{Frost}
& \multicolumn{1}{c}{Fog}
& \multicolumn{1}{c}{Brit.}
& \multicolumn{1}{c}{Contr.}
& \multicolumn{1}{c}{Elast.}
& \multicolumn{1}{c}{Pixel.}
& \multicolumn{1}{c}{JPEG}
& \multicolumn{1}{c}{} \\
\midrule
\multirow{14}{*}{\rotatebox{90}{ImageNet-C}} & Zero Shot & 11.1 & 12.5 & 12.1 & 23.3 & 15.3 & 24.4 & 22.7 & 32.4 & 29.9 & 35.9 & 54.1 & 17.3 & 12.7 & 31.0 & 33.3 & 24.5 \\
 & TDA~\cite{karmanov2024efficient} & 12.4 & 14.8 & 14.8 & 23.7 & 16.7 & 26.1 & 24.0 & 33.5 & 32.1 & 38.0 & 55.3 & 19.3 & 14.2 & 33.5 & 35.0 & 26.2 \\
 & DMN~\cite{zhang2024dual} & 11.5 & 14.1 & 14.0 & 22.3 & 15.8 & 24.6 & 22.5 & 31.5 & 30.4 & 35.3 & 54.2 & 16.3 & 12.5 & 32.5 & 24.8 & 24.2 \\
 & VTE~\cite{dobler2024lost} & 9.2 & 10.8 & 10.8 & 24.7 & 14.4 & 24.3 & 25.2 & 35.4 & 32.4 & 38.2 & 55.5 & 16.2 & 14.3 & \underline{38.7} & 34.0 & 25.6 \\
 & ZERO~\cite{farina2024frustratingly} & 10.5 & 11.5 & 11.1 & 24.8 & 14.9 & 24.4 & 23.3 & 33.0 & 30.4 & 37.1 & 54.5 & 16.9 & 13.4 & 34.9 & 32.9 & 24.9 \\
 & ECALP~\cite{li2025efficient} & 13.3 & 15.6 & 14.1 & 22.6 & 15.6 & 25.5 & 23.6 & 31.0 & 30.0 & 35.8 & 52.1 & 18.3 & 13.5 & 33.1 & 33.3 & 25.2 \\
 & TENT~\cite{wang2020tent} & 5.2 & 5.7 & 7.4 & 25.3 & 19.3 & 26.9 & 24.2 & 33.5 & 30.5 & 37.8 & 54.3 & 22.5 & 13.9 & 35.1 & 36.0 & 25.2 \\
 & RoTTA~\cite{yuan2023robust} & 11.4 & 13.1 & 12.4 & 23.5 & 15.5 & 24.7 & 22.8 & 32.6 & 29.9 & 35.9 & 54.2 & 17.3 & 12.9 & 31.0 & 33.5 & 24.7 \\
 & TPT~\cite{shu2022test} & 8.5 & 9.4 & 10.2 & 24.1 & 15.1 & 25.0 & 24.0 & 34.0 & 32.2 & 37.1 & 55.5 & 16.5 & 13.7 & 33.9 & 33.6 & 24.9 \\
 & MEMO~\cite{zhang2022memo} & 11.1 & 12.4 & 12.0 & 23.5 & 15.4 & 24.6 & 22.9 & 32.5 & 30.1 & 35.9 & 54.0 & 17.5 & 12.8 & 31.2 & 33.4 & 24.6 \\
 & WATT~\cite{osowiechi2024watt} & 11.2 & 12.8 & 13.0 & 25.1 & 18.1 & 26.9 & 25.2 & 33.4 & 30.2 & 37.4 & 54.0 & 21.8 & 15.2 & 33.5 & 34.9 & 26.2 \\
 & Mint~\cite{bao2026mint} & 19.7 & 20.4 & 19.4 & \underline{26.9} & 21.5 & 29.9 & 26.0 & 32.7 & 29.4 & 38.8 & 55.1 & 24.3 & 19.1 & 36.4 & \underline{38.2} & 29.2 \\
 & BATCLIP~\cite{maharana2025batclip} & 19.5 & 21.0 & 19.3 & 25.9 & 21.3 & 30.0 & 28.5 & 35.2 & 31.4 & 40.2 & 55.2 & 25.7 & 23.7 & 36.8 & 37.2 & 30.1 \\
 & SubTTA~\cite{zeng2026subspace} & \underline{21.2} & \underline{22.9} & \underline{21.4} & 25.3 & \underline{24.8} & \underline{30.7} & \underline{29.6} & \underline{37.2} & \underline{35.9} & \underline{41.3} & \underline{57.0} & \underline{29.7} & \underline{29.4} & 37.6 & \underline{38.2} & \underline{32.1} \\
  & \textbf{CDIM} (Ours) & \bf{26.0} & \bf{27.9} & \bf{26.7} & \bf{31.1} & \bf{30.1} & \bf{35.9} & \bf{34.4} & \bf{42.2} & \bf{38.3} & \bf{45.6} & \bf{59.6} & \bf{34.1} & \bf{33.1} & \bf{45.1} & \bf{43.8} & \bf{36.9} \\
\midrule
\multirow{14}{*}{\rotatebox{90}{CIFAR10-C}} & Zero Shot & 37.9 & 41.7 & 54.4 & 71.7 & 40.9 & 67.9 & 73.7 & 73.9 & 77.3 & 70.2 & 84.4 & 62.4 & 53.8 & 47.7 & 59.5 & 61.2 \\
 & TDA~\cite{karmanov2024efficient} & 40.9 & 44.3 & 50.5 & 72.8 & 44.4 & 71.6 & 76.0 & 75.3 & 77.8 & 71.8 & 86.1 & 62.3 & 56.3 & 47.8 & 57.7 & 62.4 \\
 & DMN~\cite{zhang2024dual} & 43.3 & 47.5 & 54.1 & 73.4 & 40.0 & 72.2 & 76.0 & 74.3 & 76.5 & 71.2 & 84.5 & 60.6 & 54.1 & 47.2 & 57.1 & 62.1 \\
 & VTE~\cite{dobler2024lost} & 42.3 & 46.2 & 64.2 & 71.1 & 45.6 & 68.5 & 73.7 & 76.7 & 78.3 & 71.1 & 85.3 & 57.2 & 59.6 & 60.6 & 61.9 & 64.2 \\
 & ZERO~\cite{farina2024frustratingly} & 39.2 & 43.8 & 57.2 & 71.9 & 40.7 & 69.2 & 74.1 & 74.4 & 77.2 & 72.4 & 83.8 & 60.7 & 55.6 & 49.0 & 62.3 & 62.1 \\
 & ECALP~\cite{li2025efficient} & 46.5 & 50.8 & 61.0 & 71.8 & 41.5 & 70.5 & 75.4 & 75.8 & 79.0 & 71.2 & 86.0 & 60.7 & 58.1 & 49.8 & 61.2 & 64.0 \\
 & TENT~\cite{wang2020tent} & 15.4 & 18.3 & 38.2 & 81.4 & 21.5 & 76.3 & 82.2 & 83.6 & 82.2 & 80.6 & 89.8 & 80.7 & 63.5 & 58.8 & 56.3 & 61.9 \\
 & RoTTA~\cite{yuan2023robust} & 39.2 & 42.6 & 55.3 & 72.1 & 41.2 & 68.1 & 74.0 & 74.4 & 78.0 & 70.8 & 84.8 & 63.1 & 54.6 & 49.4 & 60.1 & 61.9 \\
 & TPT~\cite{shu2022test} & 37.8 & 42.2 & 60.7 & 72.8 & 44.8 & 69.7 & 75.4 & 76.0 & 78.9 & 72.2 & 85.7 & 62.0 & 58.9 & 55.2 & 62.6 & 63.6 \\
 & MEMO~\cite{zhang2022memo} & 37.6 & 41.3 & 55.6 & 72.1 & 41.6 & 68.6 & 74.0 & 75.0 & 77.3 & 71.5 & 84.8 & 62.1 & 55.5 & 49.2 & 61.0 & 61.8 \\
 & WATT~\cite{osowiechi2024watt} & 46.0 & 53.2 & 60.3 & 75.0 & 38.6 & 71.5 & 75.9 & 77.7 & 80.2 & 76.3 & 87.6 & 75.6 & 55.4 & 62.0 & 63.2 & 66.6 \\
 & Mint~\cite{bao2026mint} & 54.4 & 58.7 & 64.5 & 76.5 & 48.9 & 77.9 & 79.3 & 81.6 & 81.5 & 77.2 & 89.9 & 74.5 & 62.1 & 61.4 & 64.5 & 70.2 \\
 & BATCLIP~\cite{maharana2025batclip} & \underline{61.9} & 65.4 & 67.1 & 80.1 & 55.5 & 80.0 & 81.6 & 82.5 & 83.7 & 80.4 & 88.5 & 81.2 & 69.1 & 62.7 & 67.3 & 73.8 \\
 & SubTTA~\cite{zeng2026subspace} & 60.2 & \underline{66.6} & \underline{69.7} & \underline{81.5} & \underline{59.9} & \underline{81.2} & \underline{82.4} & \underline{83.7} & \underline{83.8} & \underline{83.5} & \underline{91.4} & \underline{86.3} & \underline{70.9} & \underline{74.4} & \underline{69.4} & \underline{76.3} \\
  & \textbf{CDIM} (Ours) & \bf{67.8} & \bf{70.9} & \bf{76.3} & \bf{83.2} & \bf{67.1} & \bf{83.0} & \bf{85.2} & \bf{86.2} & \bf{87.0} & \bf{84.8} & \bf{92.5} & \bf{87.0} & \bf{74.9} & \bf{77.5} & \bf{73.4} & \bf{79.8} \\
\midrule
\multirow{14}{*}{\rotatebox{90}{CIFAR100-C}} & Zero Shot & 19.6 & 21.4 & 25.3 & 42.5 & 20.1 & 43.1 & 47.9 & 48.4 & 49.7 & 41.6 & 57.0 & 34.5 & 29.2 & 23.9 & 32.5 & 35.8 \\
 & TDA~\cite{karmanov2024efficient} & 22.6 & 25.1 & 29.2 & 43.1 & 19.4 & 43.5 & 49.3 & 48.5 & 50.4 & 41.5 & 58.0 & 35.0 & 28.7 & 24.2 & 32.5 & 36.7 \\
 & DMN~\cite{zhang2024dual} & 22.0 & 25.1 & 25.4 & 40.9 & 16.1 & 43.7 & 49.4 & 45.7 & 47.9 & 39.8 & 57.6 & 32.2 & 25.9 & 22.9 & 31.0 & 35.0 \\
 & VTE~\cite{dobler2024lost} & 18.0 & 18.8 & 28.2 & 40.4 & 19.6 & 39.6 & 45.4 & 48.2 & 46.9 & 40.7 & 55.3 & 30.1 & 32.5 & 30.4 & 31.5 & 35.0 \\
 & ZERO~\cite{farina2024frustratingly} & 18.9 & 21.1 & 28.6 & 44.0 & 19.5 & 43.3 & 48.7 & 49.1 & 50.1 & 44.0 & 57.6 & 34.5 & 31.1 & 24.8 & 34.0 & 36.6 \\
 & ECALP~\cite{li2025efficient} & 23.1 & 25.2 & 30.1 & 43.2 & 19.6 & 42.8 & 49.7 & 48.0 & 50.0 & 42.3 & 58.4 & 34.6 & 30.1 & 25.2 & 32.3 & 37.0 \\
 & TENT~\cite{wang2020tent} & 7.6 & 8.2 & 8.3 & \underline{51.7} & 8.2 & \underline{52.5} & 53.3 & 52.2 & 36.3 & 47.9 & 62.6 & \bf{52.5} & 36.4 & 40.0 & 38.1 & 37.0 \\
 & RoTTA~\cite{yuan2023robust} & 20.6 & 22.2 & 26.3 & 42.5 & 20.3 & 43.2 & 48.1 & 48.6 & 50.0 & 41.7 & 57.2 & 34.5 & 29.2 & 25.1 & 32.9 & 36.2 \\
 & TPT~\cite{shu2022test} & 17.9 & 19.5 & 27.1 & 43.5 & 20.0 & 42.6 & 48.7 & 49.1 & 49.5 & 42.2 & 57.3 & 33.4 & 31.1 & 27.6 & 32.8 & 36.1 \\
 & MEMO~\cite{zhang2022memo} & 20.2 & 21.9 & 27.2 & 44.1 & 19.6 & 43.9 & 49.1 & 50.6 & 50.4 & 43.9 & 58.5 & 34.6 & 30.4 & 25.2 & 34.2 & 36.9 \\
 & WATT~\cite{osowiechi2024watt} & 25.6 & 27.3 & 31.5 & 48.5 & 23.3 & 48.4 & 53.1 & 52.9 & 52.3 & 48.1 & 62.5 & 45.1 & 35.5 & 36.7 & 37.8 & 41.9 \\
 & Mint~\cite{bao2026mint} & 27.5 & 30.5 & 36.0 & 49.3 & 26.5 & 47.4 & 53.4 & 52.6 & 52.1 & \underline{48.5} & \underline{64.7} & 44.1 & 35.2 & 33.0 & 35.4 & 42.4 \\
 & BATCLIP~\cite{maharana2025batclip} & 25.5 & 28.4 & 34.6 & 49.7 & 26.5 & 48.8 & \underline{54.8} & 51.9 & 51.6 & 48.3 & 62.8 & 45.7 & 34.8 & 33.0 & 37.2 & 42.2 \\
 & SubTTA~\cite{zeng2026subspace} & \underline{31.9} & \underline{34.2} & \underline{36.3} & 49.5 & \underline{29.7} & 48.9 & 53.9 & \underline{53.6} & \underline{54.1} & 48.2 & 63.2 & 46.5 & \underline{36.5} & \underline{40.2} & \underline{40.8} & \underline{44.5} \\
  & \textbf{CDIM} (Ours)& \bf{33.9} & \bf{36.5} & \bf{42.8} & \bf{53.4} & \bf{32.5} & \bf{52.7} & \bf{58.0} & \bf{56.6} & \bf{56.5} & \bf{53.2} & \bf{67.4} & \underline{52.2} & \bf{41.1} & \bf{44.0} & \bf{41.7} & \bf{48.2} \\
\bottomrule
\end{tabular}
\vspace{-10pt}
\end{table*}

\newpage
\section{Pseudocode}
\label{app:pseudo}
Algorithm~\ref{alg:cdim} summarizes one \textsc{CDIM} update on a test batch of $B$ images. The inputs are the frozen DINO\,v2 backbone, CLIP with its class prompts, and the running class prior of the online logit adjustment. Only $\theta$, the set of channel-wise affine parameters of the normalization layers of CLIP, is adapted, following~\cite{wang2020tent} and~\cite{maharana2025batclip}. The DINO\,v2 backbone is frozen, requires one forward pass per batch, and receives no gradient. The update returns the adjusted predictions $\hat y_i$ and the updated EMA prior, and it updates $\theta$ in place.

\begin{algorithm}[t]
\caption{PyTorch-style pseudo-code of \textsc{CDIM} for one iteration.}
\label{alg:cdim}
\begin{lstlisting}[style=pytorchstyle]
# clip: CLIP; the affine parameters of all normalization layers
#   (image and text encoders) are trainable (optimizer)
# dino: frozen DINO v2
# prompts: tokenized class prompts
# ema_prior: running class prior, shape [K]
# rho: discard ratio (0.05), alpha: EMA momentum (0.9)
# tau_inv: inverse temperature of the DINO branch (20.0)
# logits_clip already include the pretrained CLIP logit scale
# (numerical constants such as 1e-8 and mixed-precision scaling are omitted)

def cdim_iteration(x, clip, dino, prompts, ema_prior):
    B = x.size(0)

    # 1. Forward and extract features
    # (text prototypes T are recomputed by the current CLIP at every iteration)
    logits_clip, C, T = clip(x, prompts)
    D = dino(x)

    # 2. Cross-modal logit ensemble (no gradient through the DINO branch)
    with torch.no_grad():
        Cn, Dn, Tn = [F.normalize(v, p=2, dim=1) for v in (C, D, T)]
        M = Cn.t() @ Dn / B                       # cross-covariance, [d_c, d_d]
        T_D = F.normalize(M.t() @ Tn.t(), p=2, dim=0) # column-wise L2
        logits_dino = tau_inv * (Dn @ T_D)
    Z_ens = 0.5 * logits_clip + 0.5 * logits_dino
    p = F.softmax(Z_ens, dim=1)

    # 3. Adaptive confidence gating (discard the rho*B least confident)
    E = -torch.logsumexp(Z_ens, dim=1)
    k = max(1, int(rho * B))
    G = (E < E.topk(k)[0][-1]).float()

    # 4. GT-IM objective and parameter update
    ent = -(p * p.log()).sum(1)                   # per-sample entropy
    cond = (ent * G).sum() / G.sum()              # gated conditional term
    p_bar = p.mean(0)
    marg = (p_bar * p_bar.log()).sum()            # negative marginal entropy
    loss = cond + marg
    optimizer.zero_grad(); loss.backward(); optimizer.step()

    # 5. Online logit adjustment (OLA) and prediction
    with torch.no_grad():
        ema_prior = alpha * ema_prior + (1 - alpha) * p_bar
        Z_adj = Z_ens - ema_prior.log()
        y_hat = Z_adj.argmax(1)
    return y_hat, ema_prior
\end{lstlisting}
\end{algorithm}

\end{document}